\documentclass{article}

\usepackage[final]{neurips_2024}

\usepackage[OT1]{fontenc} 
\usepackage{rotating}
\usepackage{graphicx}
\usepackage{nameref}
\usepackage[inline]{enumitem}
\usepackage{subcaption}
\usepackage{amsmath,amsthm,trimclip}
\usepackage{amssymb}
\usepackage{amsfonts}
\usepackage{footnote}
\usepackage{url}
\usepackage{lipsum}
\usepackage[inline]{enumitem}
\usepackage[acronym]{glossaries}
\usepackage{bbm}
\usepackage{xcolor}
\usepackage[hidelinks]{hyperref}
\usepackage{adjustbox}
\usepackage{algorithm} 
\usepackage{algpseudocode}
\usepackage{booktabs}
\usepackage{mdframed}
\usepackage{listings}
\usepackage{verbatim}
\usepackage{newfloat}
\usepackage{placeins}

\everypar{\looseness=-1}
\definecolor{gemma}{HTML}{1B98E0}
\definecolor{llama}{HTML}{76C7C0}
\definecolor{mistral}{HTML}{FFD79D}
\definecolor{commandr}{HTML}{02475E}

\usepackage{fancyvrb}

\newmdenv[
  backgroundcolor=gray!2,                  %
  linecolor=gray!20,                       %
  linewidth=0.5pt,                         %
  roundcorner=5pt,                         %
  font=\sffamily,                          %
  frametitlefont=\sffamily\bfseries,       %
  frametitlerule=false,                    %
  frametitlealignment=\center,             %
  innertopmargin=0cm,                      %
  innerbottommargin=1em,                   %
  skipabove=0cm,                           %
  skipbelow=0cm,                           %
]{llmbox}

\makeatletter
\def\@xobeysp{\mbox{}\space}
\def\verbatim@font{\normalfont\tiny\ttfamily\raggedright}
\makeatother

\DeclareFloatingEnvironment[
  fileext=lop,
  listname={List of Prompts},
  name=Prompt,
  placement=htp,
  within=none, %
]{prompt}

\DeclareFloatingEnvironment[
  fileext=lop,
  listname={List of Responses},
  name=Response,
  placement=htp,
  within=none, %
]{response}

\newtheoremstyle{postulate}{1em}{1em}{}{}{\bfseries}{\normalfont\textit{.}}{.5em}{}
\theoremstyle{postulate}

\author{%
Eduardo Pignatelli \\
University College London \\
\texttt{e.pignatelli@ucl.ac.uk}
\And
Johan Ferret \\
Google DeepMind \\
\And
Tim Rocktäschel \\
University College London \\\&  Google DeepMind \\
\And
Edward Grefenstette \\
University College London\\\&  Google DeepMind \\
\And
Davide Paglieri \\
University College London \\
\And
Samuel Coward \\
University of Oxford \\
\And
Laura Toni \\
University College London \\
}

\begin{document}

\title{CALM: Credit Assignment with Language Models}
\title{Assessing the Zero-Shot Capabilities of LLMs \\for Action Evaluation in RL}
\maketitle

\newacronym{rl}{RL}{Reinforcement Learning}
\newacronym{drl}{Deep RL}{Deep Reinforcement Learning}
\newacronym[longplural=Markov Decision Processes]{mdp}{MDP}{Markov Decision Process}
\newacronym[longplural=Markov Processes]{mp}{MP}{Markov Process}
\newacronym{pomdp}{POMDP}{Partially-observable MDP}
\newacronym[longplural=Markov Reward Processes]{mrp}{MRP}{Markov Reward Process}
\newacronym[longplural=Random Processes]{rp}{RP}{Random Process}
\newacronym{ca}{CA}{Credit Assignment}
\newacronym{cap}{CAP}{Credit Assignment Problem}
\newacronym{hi}{HI}{Hindsight Information}
\newacronym{gpi}{GPI}{Generalised Policy Iteration}
\newacronym{mpi}{MPI}{Modified Policy Iteration}
\newacronym{pe}{PE}{Policy Evaluation}
\newacronym{pi}{PI}{Policy Improvement}
\newacronym{lstm}{LSTM}{Long-Short-Term-Memory}
\newacronym{td}{TD}{Temporal Difference}
\newacronym{kl}{KL}{Kullback-Leibler}
\newacronym{dnn}{DNN}{Deep Neural Network}
\newacronym{llm}{LLM}{Large Language Model}
\newacronym{vlm}{LLM}{Vision Language Model}
\newacronym{calm}{CALM}{Credit Assignment with Language Models}
\newacronym{hrl}{HRL}{Hierarchical Reinforcement Learning}

\begin{abstract}
The temporal credit assignment problem is a central challenge in \gls{rl}, concerned with attributing the appropriate influence to each actions in a trajectory for their ability to achieve a goal.
However, when feedback is delayed and sparse, the learning signal is poor, and action evaluation becomes harder.
Canonical solutions, such as \textit{reward shaping} and \textit{options}, require extensive domain knowledge and manual intervention, limiting their scalability and applicability.
In this work, we lay the foundations for \gls{calm}, a novel approach that leverages \glspl{llm} to automate credit assignment via reward shaping and options discovery.
\gls{calm} uses \glspl{llm} to decompose a task into elementary subgoals and assess the achievement of these subgoals in state-action transitions.
Every time an option terminates, a subgoal is achieved, and \gls{calm} provides an auxiliary reward.
This additional reward signal can enhance the learning process when the task reward is sparse and delayed without the need for human-designed rewards.
We provide a preliminary evaluation of \gls{calm} using a dataset of human-annotated demonstrations from MiniHack, suggesting that \glspl{llm} can be effective in assigning credit in zero-shot settings, without examples or \gls{llm} fine-tuning. 
Our preliminary results indicate that the knowledge of  \glspl{llm} is a promising prior for credit assignment in RL, facilitating the transfer of human knowledge into value functions.
\end{abstract}

\section{Introduction}
The \gls{cap} \citep{minsky1961steps,sutton1984temporal,pignatelli2024a} is a fundamental challenge in \gls{rl}.
It typically involves determining the contribution of each action to the final outcome, a process crucial for accurate policy evaluation.
Effective \gls{ca} enables agents to learn useful associations between actions and outcomes, and provides useful directions to improve the policy.

However, when rewards are dispensed only at the end of a task \citep{efroni2021reinforcement}, as it is often the case, the feedback becomes sparse and delayed, making \gls{ca} particularly challenging.
In such scenarios, rewarding events are rare, and \gls{drl} agents often struggle to convert occasional successes into a robust decision-making process.
To exacerbate the issue, \gls{rl} agents typically begin with no prior knowledge (\textit{tabula rasa}) and must learn the nuances and intricacies of complex tasks from scratch.
The lack of controlled experimental conditions, such as the ability to observe counterfactuals, makes it difficult for them to distinguish between correlation and causation.
As a result, tasks that are usually easy to solve for humans become hard to address for an \gls{rl} agent.

To address these challenges, many methods incorporate prior human knowledge into RL systems.
Two techniques are canon: reward shaping \citep{ng1999policy,gupta2022unpacking} and \gls{hrl} \citep{al2015hierarchical,sutton1999between} via options \citep{sutton1999between}.
Reward shaping involves providing an additional synthetic reward to guide the agent's actions when natural rewards are uninformative.
\gls{hrl} decomposes complex tasks into simpler ones (\textit{options}), training agents to achieve intermediate objectives that provide a signal while the \gls{mdp} would not.
Despite their effectiveness, these methods require extensive human input, making them costly and difficult to scale across different environments.

Recently, \glspl{llm} have emerged as a useful tool to transfer human knowledge into computational agents, either through planning \citep{dalal2024planseqlearn}, expressing preferences \citep{klissarov2023motif}, or grounding their abstract knowledge into practical solutions \citep{huang2023grounded,carta2023grounding}.
Notably, these models have produced strong results in causal reasoning tasks \citep{jin2023cladder} with performances comparable to humans \citep{kiciman2023causal}.
These results suggest that \glspl{llm} could be an effective, supplementary tool to distinguish between correlation and causation more effectively than traditional methods used in early stages of RL training.

With these results, a natural question arises: ``\textit{Can the knowledge encoded in \glspl{llm} serve as a useful prior for \gls{ca} in \gls{rl}?}''
Inspired by the successes of \glspl{llm}, we introduce \gls{calm},  a general method to perform \gls{ca} with \glspl{llm} using reward shaping.
We hypothesize that the prior knowledge of a \gls{llm} can provide valuable signals that improve \gls{ca} in \gls{rl}, and propose a way to transfer these priors into the agent's value function.
On this assumption, \gls{calm} leverages a pretrained \gls{llm} to break down tasks into smaller, composable subgoals and determine if a state-action-state transition achieves a subgoal.
This provides an additional reward signal to enhance RL algorithms, and effectively automates reward shaping by substantially reducing the involvement of humans in the training loop.

We present a preliminary evaluation of the efficacy of \gls{calm} in zero-shot settings, with no examples and no finetuning.
We collect a dataset of demonstrations from MiniHack \citep{samvelyan2021minihack} and use it to compare the performance of \glspl{llm} against human annotations.
Our results indicate that \glspl{llm} are a viable means to transfer common human knowledge into value functions, and can be effective in automating reward shaping.
This bodes well for the prospect to improve \gls{ca} in the full \gls{rl} problem.

\section{Related work}
\paragraph{\glspl{llm} for \gls{rl}.}
Recent advancements have shown the potential of pretrained \glspl{llm} in enhancing \gls{rl} agents.
\citet{paischer2022history,paischer2024semantic} used CLIP encodings to improve the state representations of \glspl{pomdp}.
\citet{yao2020keep,du2023guiding} investigated the ability of pretrained \glspl{llm} to improve exploration.
\citet{huang2023grounded,carta2023grounding} grounded the abstract knowledge of these models and their capabilities into practical \gls{rl} tasks.
\glspl{llm} have been used for planning, either directly as world models \citep{huang2022language,wang2023describe,singh2023progprompt,brohan2023can,dasgupta2023collaborating,shah2023lm,zhong2020rtfm,zhong2022improving} or by writing code \citep{liang2022code}.
Unlike these methods we use pretrained \glspl{llm} as a critic: the \gls{llm} provides an \emph{evaluation} of an action for how useful it is to achieve a goal in the future.
Among the methods above, \citet{du2023guiding} is the only method to use subgoals, but these are used to condition a goal-oriented policy, rather than as a critic.

\paragraph{LLMs for reward shaping.}
\citet{carta2022eager, goyal2019using} explore the advantages of using pure language abstractions for reward shaping, but do not use a pretrained \glspl{llm} and its prior knowledge.
\citet{kwon2023reward} use the responses of \glspl{llm} as a reward signal, but the investigation is limited to conversational environments.

\paragraph{LLMs for knowledge transfer.}
Another set of studies used intrinsic rewards to transfer the prior knowledge of an \gls{llm} to a value function.
\citet{wu2024read} used \glspl{llm} to provide an auxiliary reward signal in Atari \citep{bellemare2013arcade}, based on the information contained in a game manual.
Unlike this study, we use subgoals to extract the reward signal, and we do not focus on incorporating external knowledge material, but rely on the \gls{llm}'s prior knowledge to solve the task.
\citet{klissarov2023motif} constructed a reward function from the \gls{llm}'s preferences over NetHack \citep{kuttler2020nethack} in-game messages only.
Instead, our method incorporates the full observation, does not use preferences, and does not require a separate stage to fit the preference set, but uses the \gls{llm}'s output directly. \looseness=-1

In short, none of these methods proposes to generalise reward shaping with hierarchical skills using pretrained \glspl{llm}.
Unlike the methods above, we use pretrained \glspl{llm} as a critic: we aim to uncover cause-effect relationships between actions and goals by both breaking down a task into valuable subgoals and then acting as a reward function for them.
This provides an intermediate signal to shape the agent's behaviour when rewards are sparse and delayed.

\section{Preliminaries}
We consider the problem of learning to solve \glspl{pomdp}.
A \gls{pomdp} is defined by a tuple $\mathcal{M} = (\mathcal{S}, \mathcal{A}, R, \mu, \mathcal{O}, O, \gamma)$.
Here $\mathcal{S}$ is the state space with elements $s$.
$\mathcal{A}$ is the action space of elements $a$.
$R: \mathcal{S} \times \mathcal{A} \times \mathcal{S} \rightarrow [0, 1]$ is a deterministic, bounded reward function.
$\mu: \mathcal{S} \times \mathcal{A} \rightarrow \mathcal{S}$ is the state transition function.
$\mathcal{O}$ is the space of all observations, and $O: \mathcal{S} \rightarrow \mathcal{O}$ is an observation function, mapping a state $s$ to a partial observation $o$.
$\gamma \in [0, 1]$ is the discount factor.

To best isolate the \gls{cap} from other problems, we focus only on environments with a discrete action space, and deterministic state transitions.
To evaluate the capabilities of \glspl{llm} in environments where the \gls{cap} is hard, we only consider tasks where the reward signal is delayed.
Here, the reward function is $0$ everywhere, and $1$ when a goal state is reached.

To start the investigation, we evaluate the \gls{llm} only in language settings, and do not consider multimodal (text, image, audio, video) settings.
For this reason, we consider only environments with an observation kernel that maps to a textual codomain, $O: \mathcal{S} \rightarrow \mathcal{T}$,  where $\mathcal{T}$ is a set of sequences of characters.

Finally, we consider a black box, pretrained \gls{llm}, that takes an input text and maps it to a finite set of output characters.
We consider only open-weights models that can fit an NVIDIA A100 80Gb in either 16 bits floating point or 4 bits integer mode.
We assume that the \gls{llm} has enough preliminary knowledge of the MiniHack environment to recognise valuable actions that progress towards a win.

\section{Methods}
We set out to design a general method to assign credit in \gls{rl} using \glspl{llm} that can generalise to multiple tasks with little human input.
Next, we formalise the method, discuss its assumptions and provide details on the protocols we use to evaluate it.

\subsection{Reward shaping}
Among the available \gls{ca} techniques, we focus on \textit{reward shaping} \citep{ng1999policy}, due to both its effectiveness in assigning credit and its limitations to generalisation related to the costs of human involvement in the training loop.
Reward shaping aims to address the scarcity of learning signals by introducing an auxiliary reward function, the \textit{shaping function}:
\begin{align}
    \tilde{r}_{t+1} = \tilde{R}(s_t, a_t, s_{t+1}).
\end{align}
Here, $s_t$ is the state at time $t$, $a_t$ is the action taken in that state, $s_{t+1}$ is the resulting state, and $\tilde{r}_{t+1}$ is the auxiliary reward collected after taking $a_t$ in $s_t$.
This reward is added to the original reward signal $R(s_t, a_t, s_{t+1})$ to obtain the new, shaped reward
\begin{align}
    r_{t+1} = R(s_t, a_t, s_{t+1}) + \tilde{R}(s_t, a_t, s_{t+1}).
\end{align}
If there exist a function $\phi: \mathcal{S} \rightarrow \mathbb{R}$ such that $\tilde{R}(s_t, a_t, s_{t+1}) = \phi(s_{t+1}) - \phi(s_t)$, then the set of optimal policies is preserved, and the shaping function is also a \textit{potential function} \citep{ng1999policy}.
In the following, we consider the more general case of non-optimality preserving functions.

For example, in key-door environments, a common testbed for \gls{ca} methods \citep{hung2019optimizing,mesnard2021counterfactual}, the agent must reach a reward behind a locked door, which can only be opened if the agent possesses a key.
Here, the agent has clear subgoals:
\begin{enumerate*}[label=(\textit{\roman*})]
    \item go to the key,
    \item pick it up,
    \item go to the door,
    \item unlock it,
    \item go to the reward.
\end{enumerate*}
Achieving these subgoals sequentially leads to optimal behaviour.
However, the agent struggles to recognise this hierarchical pattern due to the lack of immediate feedback from the environment.
This is particularly true in the early stages of training, when behaviour is erratic, and two optimal actions can be separated by a long sequence of random ones.
Providing intermediate feedback for each achievement often improves the agent's performance \citep{gupta2022unpacking}, and the ability of $\tilde{R}$ to produce an instantaneous signal indicating progress is crucial for better \gls{ca}.
Thus, reward shaping can significantly accelerate the learning process in environments with sparse or delayed rewards.

However, designing an effective shaping function is challenging.
The function should be carefully designed to provide useful guidance without leading to unintended behaviours.
This often calls for incorporating domain knowledge or heuristic information about the task, and requires deep task and environment knowledge.
Such knowledge may not be readily available or easily codifiable, limiting the applicability of reward shaping in diverse or unknown environments.
This process is complex and time-consuming, and it might not always be possible to devise a reward function that incentivizes learning, is computationally cheap, and general enough to adapt to various tasks.
Improving this limitation could enable broader use of reward shaping and enhance \gls{ca} in deep RL. \looseness=-1

\subsection{LLMs as shaping functions}
Encouraged by the recent successes of \glspl{llm} in \gls{rl} \citep{klissarov2023motif} and of using language to abstract skills \citep{jiang2019language, jacob2021multitasking, sharma2021skill,mu2022improving}, we explore whether these models can offer a valid alternative to humans in the reward shaping process.
Our goal is to produce a function that, given a description of the task and a state-action-state transition, produces a binary signal indicating whether the action makes progress towards solving the task or not: \looseness=-1
\begin{align}
    \textrm{LLM}: \textrm{desc}(\mathcal{M}) \times \textrm{desc}(\mathcal{S} \times \mathcal{A} \times \mathcal{S}) \rightarrow \mathbb{B}.
    \label{eq:llm}
\end{align}
Here, $\textrm{LLM}$ is a pretrained \gls{llm}; $\textrm{desc}(\mathcal{M})$ is a natural language description of the \gls{pomdp} (the task); $\textrm{desc}(\mathcal{S} \times \mathcal{A} \times \mathcal{S})$ is a textual representation of the transition, not necessarily in natural language (for example, a grid-arranged text), and $\mathbb{B} = \{0, 1\}$ is the Boolean domain.
In this scenario, the \gls{llm} acts as a critic: its role is to \textit{evaluate} the action $a_t$ in the transition $(s_t, a_t, s_{t+1})$ based on the heuristics that we describe next.

We operationalise the idea using the notion of \textit{options} \citep{sutton1999between}.
An \textit{option} is a temporally extended action and consists of two elements: an intra-option policy $\pi_i: \mathcal{S} \rightarrow \Delta(\mathcal{A})$, and a termination condition $\beta: \mathcal{S} \rightarrow \mathbb{B}$.\footnote{We consider $\mathcal{S}^+ = \mathcal{S}$ and omit the initiation set $\mathcal{S}^+$}

To develop an intuition of options, it is useful to visualise one as a macro-action: a set of actions that, taken together, have precise semantics.
For example, in our key-to-door example, one useful option to consider is to \textit{pick up the key}.
This macro action includes a set of primitive actions -- the set of actions to navigate to the key and the action \textit{pickup} -- and a termination condition -- whether the key is picked up.
For the purpose of our analysis, this termination is crucial, as it signals that the subtask has been successfully achieved.

We exploit this idea to build our shaping function, set up a single-turn conversation, and prompt the model to perform two subtasks:
\begin{enumerate}[label=\textit{(\roman*)}]
    \item To identify a set of useful options in the environment, by breaking down the task into a sequence of shorter subgoals. These options, and more specifically their termination, effectively constitutes our set of subgoals, since a subgoal is achieved when the option terminates (a key is picked up).
    \item Determine whether an option terminated (thus, if a subgoal is achieved) in the transition $(s_t, a_t, s_{t+1})$.
\end{enumerate}
Every time an option terminates, we augment the task reward with the subtask reward as according to our reward shaping rule, $\tilde{R}(s_t, a_t, s_{t+1}) = \beta(s_{t+1})$.

In essence, Equation~\eqref{eq:llm} aims to mimic a human supervising an \gls{rl} agent's decisions, acting as an auxiliary critic.
Decomposing the task into multiple subgoals allows each sub-achievement to correspond to a small step towards success, and composing all the subgoals sequentially results in successful behaviour.
Since achieving a subgoal is contingent on achieving all the preceding ones, the number of subgoals achieved quantifies the agent's progresses.
To develop an intuition of the idea, subgoals can be thought of as levels; gaining a level at the current time indicates progress in achieving a specific goal in the future.
This process of \textit{actualisation}, where an action is evaluated for its future potential to achieve a goal, characterises the function as a \gls{ca} method \citep{pignatelli2024a}.
\looseness=-1

\subsection{Experimental protocol}
The viability of \gls{calm} in online \gls{rl} settings depends on the quality of the assignments provided by the \gls{llm}.
Good quality assignments -- signals that reinforce optimal actions -- can improve the performance of an \gls{rl} algorithm.
Thus, we provide a preliminary evaluation of \gls{calm} on an offline dataset of demonstrations.

\paragraph{Environment.} We focus on the KeyRoom environment, a canonical testbed for \gls{ca} methods \citep{hung2019optimizing,mesnard2021counterfactual,mesnard2023quantile} originally proposed in Minigrid \citep{chevalier2018gym_minigrid}.
We choose its MiniHack version, for it provides a textual representation of the observations that can be fed to a language system.
The game presents a top-down view of a grid-like environment with two rooms.
The agent starts in the first room, where a key is located.
It must pick up the key and use it to unlock the door to the second room, where a reward is located.
We consider two types of observations:
\begin{enumerate}
    \item \textbf{Cropped observations.} a top-down, north-facing, $9$x$9$ crop around the agent, which is known to improve the performance in standard \gls{rl} benchmarks on Nethack \citep{kuttler2020nethack}.
    \item \textbf{Game screens.} A top-down, north-facing, $21$x$79$ grid showing the entire game scene, including an in-game message and a set of statistics of the current state. We also refer to these as \textit{human} observations, since they reproduce the conditions of human game play.
\end{enumerate}
Both observations are partial, despite containing different amounts of information.
We consider a discrete action set: \textit{go north}, \textit{go east}, \textit{go south}, \textit{go west}, \textit{pickup}, \textit{apply}.
The reward function is deterministic, providing a reward of $1$ if the agent reaches the goal tile and $0$ otherwise.
Transitions are also deterministic.

\paragraph{Dataset.} We collect 256 one-step transitions $d_t = (s_t, a_t, s_{t+1})$ using a random policy.
Given a set of subgoals $\mathcal{G} \subset (\mathcal{S} \times \mathcal{A} \times \mathcal{S})$, a transition $d_t$ can then be classified as either achieving a subgoal $g \in \mathcal{G}$ or not.
This produces categories $\mathcal{C} = \{c_i \,:\, 0 \leq i \leq |\mathcal{G}| + 1\}$, one for each subgoal, and an additional one when no subgoal is achieved.
To characterise the abilities of an \gls{llm} to assign credit accurately, we produce a balanced dataset where each goal appears with equal probability.

\paragraph{Composing the prompt.}
For each transition we then compose a prompt using the following structure:
\begin{enumerate}
    \item
    \texttt{<ROLE>} specifies the role we ask the \gls{llm} to simulate.
    \item
    \texttt{<ENVIRONMENT-DESCRIPTION>} describes the \gls{rl} environment, the source of the gameplay.
    \item
    \texttt{<SYMSET>} is a list reporting Nethack wiki entries\footnote{https://nethackwiki.com/wiki/Symset} of what each symbols in the grid represents.
    \item
    \texttt{<TASK-DESCRIPTION>} specifies the overall goal of the agent, and does not contain information about subgoals.
    \item
    \texttt{<SUBGOALS>} contains either a list of subgoals to achieve, or asks the \gls{llm} to produce one.
    \item
    \texttt{<INSTRUCTIONS>} tasks the agent to determine whether a subgoal is achieved in the trajectory presented in \texttt{<TRANSITION>}.
    \item
    Finally, \texttt{<OUTPUT-FORMAT-REQUEST>} requests the output in a format that can be easily parsed, for example, a python dictionary.
\end{enumerate}
Prompt~\ref{prompt:example} shows a concrete instantiation of this structure, where goals are provided as part of the input.
Here, the role is not specified, exhorting the \gls{llm} to play a generic role, and the environment description (\textit{The environment is MiniHack}) is minimal.
In the symset -- the list of symbols with their meaning -- the descriptions are extracted from the wiki (\url{https://nethackwiki.com/wiki/Symset}).
The task is as generic as possible (\textit{to win the game}), and it is followed by the set of predetermined subgoals (\textit{pick up the key} and \textit{unlock the door}).
The instructions and the request for an appropriate output format follow on that.
Finally, we enclose the transition within a \texttt{<gameplay>} tag, and remark that this is a single-turn conversation to avoid the model asking additional clarifications.
Notice that we separate each cell in the observation with a whitespace to ensure that each cell (plus their whitespace) corresponds to a separate token.
We discuss this more in depth in Appendix~\ref{app:ablations:tokens}, and provide more details and variations of prompts in Appendix~\ref{app:prompts}.
To develop an intuition of the role covered by the model, we encourage the reader to scan over them before proceeding.

\begin{prompt}[t]
\begin{llmbox}[frametitle=Example prompt]
\begin{verbatim}
The environment is MiniHack.

I will present you with a short extract of a gameplay. At each timestep, symbols represent the following items:
- "." represents a floor tile.
- "|" can represent either a wall, a vertical wall, an open door.
- "-" can represent either the bottom left corner (of a room), bottom right corner (of a room), wall, horizontal wall, wall, top left corner (of a room), op right corner (of a room).
- "+" represents a closed door. Doors can be locked, and require a key to open.
- "(" represents a useful item (pick-axe, key, lamp...)
- "<" represents a ladder or staircase up.
- ">" represents a ladder or staircase down.

The task of the agent is to win the game.

First, based on your knowledge of NetHack, break down the task of the agent into subgoals. 
Then, consider the following game transition, which might or might not contain these subgoals.
Determine if any of the subgoals is achieved at Time: 1 or not.


Report your response in a dictionary containing the name of the subgoals as keys and booleans as value. For example:
```python
{
    <name of goal>: <bool>,
} 

Observation Sequence:

<gameplay>
Time: 0
Current message: 

      - - - -    
      | . . |    
      | . . |    
- - + - . < |    
| . . . @ . |    
| . ( . . . |    
- - - - - - -    
                 
                 
Time: 1
Current message: 

      | . . |    
      | . . |    
- - + - . < |    
| . . . . . |    
| . ( . @ . |    
- - - - - - -    
                 
                 
                 
</gameplay>

I will not consider anything that is not in the dictionary.
You have only one shot at this, and you cannot ask for clarifications.

\end{verbatim}
\end{llmbox}
\caption{Example of a prompt for instruction verification. Here, goals are provided externally from a human.}
\label{prompt:example}
\end{prompt}

\paragraph{Models.} We use pretrained, open-weights large language models that can fit a 80Gb A100 Nvidia GPU in either 16 bits brain floating point \citep{dean2012large} or 4 bits integer weights representations.
When models cannot fit in memory, we use their NF4 \citep{dettmers2023qlora} quantised equivalent.
These models are marked with an asterisk ($^*$) in the tables below.
All the models are finetuned for instructions following, and tokens are deterministically sampled using a greedy policy.

\paragraph{Annotations.}
For each transition a human annotator produces a term of reference for comparison.
The annotator is presented with each prompt in the dataset, without any further instructions.
We then record the annotator's answer, and use it as a term of reference for the \gls{llm}'s responses.
Since the prompt has a correct answer, these are not subjective evaluations, but rather direct verification, with little room for interpretation.

\paragraph{Metrics.}
We then compare the \gls{llm}'s annotations with the human ones.
The response is a true positive if both the \gls{llm} and the human annotator identify that a subgoal is achieved.
It is a false positive (a \textit{hallucination}) if the \gls{llm} identified it, but the human has not; a false negative (a \textit{miss}) if the human identified one, but the agent has not.
This effectively casts the problem as classification, with the set of classes $\mathcal{C}$, as described in the dataset description.
We then compare the \gls{llm}'s hypotheses with the human responses as ground truth, and report accuracy, F1 score, precision and recall. \looseness=-1

\section{Experiments, results, and discussion}
\label{sec:results}
To evaluate the effectiveness of \glspl{llm} in \gls{ca} for \gls{rl}, we consider environments with textual representations.
We assume that the \gls{llm} has sufficient knowledge of the game to evaluate actions.
While this assumption might be strong for NetHack, it is reasonable for MiniHack, where tasks are simplified yet challenging models of common NetHack scenarios, requiring only partial knowledge.

Based on the set of experimental conditions described above, we then consider a spectrum of settings requiring progressively less input from humans.
We start by providing the \gls{llm} with:
\begin{enumerate*}[label=\textit{(\alph*)}]
    \item cropped observations focused around the agent;
    \item an effective, predetermined set of subgoals;
\end{enumerate*}
We then proceed to progressively relax these conditions to:
\begin{enumerate*}[label=\textit{(\alph*)}]
    \item gamescreen observations;
    \item allowing the \gls{llm} to discover useful subgoals autonomously.
\end{enumerate*}
These conditions are set to replicate the conditions of a human playing the game.

\subsection{Can LLMs understand goal specifications and verify option termination?}
\label{sec:res:preset}
This experiment aims to assess whether a pretrained \gls{llm} can function as a reward function when subgoals are provided externally.
We provide the LLM with the environment name, \texttt{MiniHack}, and a list of two subgoals: \texttt{pick up the key} and \texttt{unlock the door}.
We specify that the goal of the agent is simply to \textit{win the game} \citep{jeurissen2024playing}, and ask it to determine if each subgoal has been achieved in the transition.
Prompt~\ref{app:prompt:termination} shows an example prompt for this experiment.

We present results for multiple pretrained \glspl{llm}, using both cropped observations and full game screens.
The purpose of the comparison is not to determine a winning model.
It is, instead, to understand whether the ability to assign credit to single transitions is in the spectrum of capabilities of existing open-weights \glspl{llm}.
This will lay the foundation for applying the method in full \gls{rl} settings. \looseness=-1

\begin{table}[h]
\centering
\begin{adjustbox}{max width=1\textwidth,center}
\begin{tabular}{lrrrrrrrr}
Annotator & F1 $\uparrow$ & Accuracy $\uparrow$ & Precision $\uparrow$ & Recall $\uparrow$ & TP $\uparrow$ & TN $\uparrow$ & FP $\downarrow$ & FN $\downarrow$ \\
\midrule
Human & \textbf{1.00} & \textbf{1.00} & \textbf{1.00} & \textbf{1.00} & \textbf{171} & \textbf{85} & \textbf{0} & \textbf{0} \\
\midrule
Mixtral-8x7B-Instruct-v0.1$^*$ & \textbf{0.74} & 0.67 & 0.77 & \textbf{0.73} & \textbf{124} & 47 & 38 & \textbf{47} \\
gemma-1.1-7b-it & 0.73 & \textbf{0.70} & 0.91 & 0.61 & 105 & 75 & 10 & 66 \\
Meta-Llama-3-70B-Instruct & 0.66 & 0.65 & \textbf{0.97} & 0.50 & 85 & 82 & 3 & 86 \\
Meta-Llama-3-8B-Instruct & 0.64 & 0.64 & 0.95 & 0.49 & 83 & 81 & 4 & 88 \\
c4ai-command-r-v01$^*$ & 0.60 & 0.57 & 0.80 & 0.49 & 83 & 64 & 21 & 88 \\
Mistral-7B-Instruct-v0.2 & 0.48 & 0.54 & 0.96 & 0.32 & 55 & 83 & 2 & 116 \\
gemma-1.1-2b-it & 0.00 & 0.33 & 0.00 & 0.00 & 0 & \textbf{85} & \textbf{0} & 171 \\
\midrule
Random & 0.33 & 0.33 & 0.33 & 0.33 \\
\end{tabular}
\end{adjustbox}
\caption{
Performance of \gls{llm} annotations against human annotations with \textbf{game screen} observations and with the subgoals \textbf{provided} in the prompt.
Models marked with an asterisk ($^*$) are quantised to NF4 format.
TP stands for \textit{true positives}, TN for \textit{true negatives}, FP for \textit{false positives}, and FN for \textit{false negatives}.
Rows sorted by F1 score.
}
\label{tab:res:human-preset-balanced}
\end{table}

\begin{table}[h]
\centering
\begin{adjustbox}{max width=1\textwidth,center}
\begin{tabular}{lrrrrrrrr}
Annotator & F1 $\uparrow$ & Accuracy $\uparrow$ & Precision $\uparrow$ & Recall $\uparrow$ & TP $\uparrow$ & TN $\uparrow$ & FP $\downarrow$ & FN $\downarrow$ \\
\midrule
Human & \textbf{1.00} & \textbf{1.00} & \textbf{1.00} & \textbf{1.00} & \textbf{171} & \textbf{85} & \textbf{0} & \textbf{0} \\
\midrule
Mixtral-8x7B-Instruct-v0.1$^*$ & \textbf{0.78} & \textbf{0.70} & 0.78 & \textbf{0.77} & \textbf{132} & 48 & 37 & \textbf{39} \\
gemma-1.1-7b-it & 0.76 & 0.69 & 0.79 & 0.73 & 124 & 52 & 33 & 47 \\
gemma-1.1-2b-it & 0.76 & 0.68 & 0.76 & \textbf{0.77} & 131 & 43 & 42 & 40 \\
c4ai-command-r-v01$^*$ & 0.75 & 0.69 & 0.81 & 0.70 & 120 & 57 & 28 & 51 \\
Meta-Llama-3-70B-Instruct & 0.63 & 0.58 & 0.76 & 0.54 & 92 & 56 & 29 & 79 \\
Meta-Llama-3-8B-Instruct & 0.61 & 0.61 & 0.92 & 0.46 & 79 & 78 & 7 & 92 \\
Mistral-7B-Instruct-v0.2 & 0.61 & 0.62 & \textbf{0.96} & 0.45 & 77 & \textbf{82} & \textbf{3} & 94 \\
\midrule
Random & 0.33 & 0.33 & 0.33 & 0.33 \\
\end{tabular}
\end{adjustbox}
\caption{
Performance with \textbf{cropped} observations and with the subgoals \textbf{provided} in the prompt.
}
\vspace{-1em}
\label{tab:res:crop-preset-balanced}
\end{table}

We report results in Tables~\ref{tab:res:human-preset-balanced}~and~\ref{tab:res:crop-preset-balanced}, and draw the following two insights.
First, \glspl{llm}, except \textit{gemma-1.1-2b-it}, probably due to its small size, are generally effective in recognising when an instruction has been successfully completed in a state-action-state transition.
This shows their ability to understand goal specifications and to recognise when an option terminates due to completion.
We also noticed that \textit{c4ai-command-r-plus} degenerates into outputting \textit{false} for most transitions, most probably due to quantisation.

Second, restricting the field of view of the observation helps improve performance.
This is most likely due to observations being more concise, and avoiding the information to drown among a high number of tokens.
This also seems to increase the lower bound, and the performance of models drastically failing with \textit{human} observations greatly improves, especially \textit{gemma-1.1-2b-it}.

\subsubsection{Can LLMs suggest effective options?}
In this experiment, we evaluate whether \glspl{llm} can autonomously suggest effective options.
Instead of providing a predetermined list, we ask the LLM to break down the task into subgoals and verify whether these subgoals have been achieved.
Despite only a small change on the surface, removing some key information from the prompt intensively tests the \gls{llm}'s knowledge of NetHack.
More importantly, it stresses the ability of the models to come up with a viable and effective hierarchy of subgoals such that, if reinforced, produces useful signals for progress.

This setting is more complex but also more general, as it replicates the amount of information typically available to a human player.
Prompt~\ref{app:prompt:discovery} shows an example prompt for this experiment.
As for the previous experiment, we evaluate the performance of different models using both cropped and human observations.
\begin{table}[h]
\centering
\begin{adjustbox}{max width=1\textwidth,center}
\begin{tabular}{lrrrrrrrr}
Annotator & F1 $\uparrow$ & Accuracy $\uparrow$ & Precision $\uparrow$ & Recall $\uparrow$ & TP $\uparrow$ & TN $\uparrow$ & FP $\downarrow$ & FN $\downarrow$ \\
\midrule
Human & \textbf{1.00} & \textbf{1.00} & \textbf{1.00} & \textbf{1.00} & \textbf{171} & \textbf{85} & \textbf{0} & \textbf{0} \\
\midrule
Meta-Llama-3-70B-Instruct & \textbf{0.82} & \textbf{0.72} & 0.71 & \textbf{0.96} & \textbf{165} & 19 & 66 & \textbf{6} \\
Meta-Llama-3-8B-Instruct & 0.80 & 0.70 & 0.72 & 0.89 & 153 & 26 & 59 & 18 \\
gemma-1.1-7b-it & 0.77 & 0.66 & 0.71 & 0.85 & 145 & 25 & 60 & 26 \\
Mixtral-8x7B-Instruct-v0.1$^*$ & 0.74 & 0.64 & 0.71 & 0.76 & 130 & 33 & 52 & 41 \\
Mistral-7B-Instruct-v0.2 & 0.57 & 0.48 & 0.63 & 0.53 & 90 & 32 & 53 & 81 \\
c4ai-command-r-v01$^*$ & 0.56 & 0.52 & 0.71 & 0.47 & 80 & 52 & 33 & 91 \\
gemma-1.1-2b-it & 0.00 & 0.33 & 0.00 & 0.00 & 0 & \textbf{85} & \textbf{0} & 171 \\
\midrule
Random & 0.33 & 0.33 & 0.33 & 0.33 \\
\end{tabular}
\end{adjustbox}
\caption{
Performance with \textbf{game screen} observations and with \textbf{autonomously discovered} subgoals.
}
\vspace{-1em}
\label{tab:res:human-suggested-balanced}
\end{table}

\begin{table}[h]
\centering
\begin{adjustbox}{max width=1\textwidth,center}
\begin{tabular}{lrrrrrrrr}
Annotator & F1 $\uparrow$ & Accuracy $\uparrow$ & Precision $\uparrow$ & Recall $\uparrow$ & TP $\uparrow$ & TN $\uparrow$ & FP $\downarrow$ & FN $\downarrow$ \\
\midrule
Human & \textbf{1.00} & \textbf{1.00} & \textbf{1.00} & \textbf{1.00} & \textbf{171} & \textbf{85} & \textbf{0} & \textbf{0} \\
\midrule
Meta-Llama-3-70B-Instruct & \textbf{0.83} & \textbf{0.75} & 0.75 & 0.93 & 159 & 33 & 52 & 12 \\
gemma-1.1-7b-it & 0.81 & 0.70 & 0.71 & \textbf{0.95} & \textbf{163} & 17 & 68 & \textbf{8} \\
Mixtral-8x7B-Instruct-v0.1$^*$ & 0.72 & 0.62 & 0.71 & 0.74 & 127 & 32 & 53 & 44 \\
Mistral-7B-Instruct-v0.2 & 0.65 & 0.54 & 0.66 & 0.64 & 109 & 28 & 57 & 62 \\
c4ai-command-r-v01$^*$ & 0.60 & 0.52 & 0.68 & 0.54 & 92 & 41 & 44 & 79 \\
gemma-1.1-2b-it & 0.47 & 0.52 & \textbf{0.89} & 0.32 & 55 & \textbf{78} & \textbf{7} & 116 \\
Meta-Llama-3-8B-Instruct & 0.45 & 0.39 & 0.57 & 0.37 & 63 & 38 & 47 & 108 \\
\midrule
Random & 0.33 & 0.33 & 0.33 & 0.33 \\
\end{tabular}
\end{adjustbox}
\caption{
Performance with \textbf{cropped} observations and with \textbf{autonomously discovered} subgoals.
}
\vspace{-2em}
\label{tab:res:crop-suggested-balanced}
\end{table}

Results in Table~\ref{tab:res:human-suggested-balanced} indicate that \glspl{llm} can effectively suggest subgoals when presented with game screen observations, and that these subgoals align with those identified by humans.
Models like \textit{Meta-Llama-3-70B-Instruct} and \textit{Meta-Llama-3-8B-Instruct} come close to human performance, suggesting that LLMs can effectively use the additional information to suggest and validate subgoals.
These results bode well for applications of \gls{calm} where human input, while still considerably smaller than in canonical reward shaping, is still expensive to collect.

When transitioning to cropped observations (Table~\ref{tab:res:crop-suggested-balanced}) \glspl{llm} perform worse.
This is most likely due to a misalignment between the subgoals proposed by the models and the ones of the ground truth.
We did not observe any substantial difference in how different models propose subgoals and in the types of subgoals they suggest.
Most models correctly identify \textit{item collection} and \textit{locating objects}, such as stairs, monsters and keys.
They often include ``going to $<$object$>$'' instructions as subgoals.
We provide examples of such prompts in Appendix~\ref{app:responses-subgoals}.

While this evaluation can be unfair, since we compare the \gls{llm}'s response with the set of subgoals the human identified, it still tells whether the \gls{llm} way of reasoning about a task align with the human one.
These elements, together with the ability of \glspl{llm} to verify if a subgoal is achieved, suggest that \glspl{llm} can be an effective means to transfer human knowledge into value functions.

\subsection{Conclusions, limitations, and future work}
In this study, we explored whether \glspl{llm} can be a useful means to transfer human knowledge into the value function of \gls{rl} agents.
By focusing on reward shaping, we highlighted its limitations in scalability due to the cost of human involvement.
To mitigate these costs, we proposed replacing humans with \glspl{llm}, leveraging their ability to decompose tasks into shorter subgoals.
Preliminary results from an offline dataset of MiniHack demonstrations suggest that \glspl{llm} are effective in verifying subgoal achievement and align with those proposed by humans.
This suggests the potential of using \glspl{llm} to enhance \gls{ca} in \gls{rl}.

\paragraph{Limitations.}
While preliminary results are promising, they are limited by the scope of the current evidence.
We did not conduct \gls{rl} experiments to validate the method in online \gls{rl} settings. 
The dynamic nature of online \gls{rl} could pose unique challenges not present in offline settings.
Additionally, despite KeyRoom being representative of the \gls{ca} challenges, and a common testbed for \gls{ca}, evaluating the method in a broader range of environments would provide more comprehensive evidence of its robustness and applicability.

The method also has inherent limitations.
Environments must provide observations in the form of text.
The \gls{llm} must hold enough knowledge of the game to evaluate actions.
While this can be a mild assumption for MiniHack, it can be an obstacle for environments requiring more specialised knowledge, such as Nethack \citep{kuttler2020nethack} or Crafter \citep{hafner2021crafter,matthews2024craftax}.
Finally, the \gls{llm} relies solely on their prior knowledge and does not incorporate new knowledge while assigning credit, limiting their adaptability and accuracy over time.
\looseness=-1

\paragraph{Future work.}
Future work should focus on addressing these limitations.
Validating the approach in online RL settings and exploring its applicability to a broader range of environments can tell if \gls{calm} can enhance the learning process of RL agents in practice.
A natural extension of this work is to generalise the method beyond text-only observations.
\citet{baumli2023vision} follows this line of research, testing the capability of \glspl{vlm} to evaluate the completion of an instruction from pixels alone.
The instruction completion question corresponds to ours in the \glspl{vlm} domain.
Finally, a closed feedback loop where \gls{calm} helps improve the policy, the policy provides new information to the \gls{llm}, and the \gls{llm} incorporates this information to improve its \gls{ca} ability could help scale to more complex problems requiring specialistic knowledge.

\bibliography{references}

\begin{thebibliography}{48}
\providecommand{\natexlab}[1]{#1}
\providecommand{\url}[1]{\texttt{#1}}
\expandafter\ifx\csname urlstyle\endcsname\relax
  \providecommand{\doi}[1]{doi: #1}\else
  \providecommand{\doi}{doi: \begingroup \urlstyle{rm}\Url}\fi

\bibitem[Minsky(1961)]{minsky1961steps}
Marvin Minsky.
\newblock Steps toward artificial intelligence.
\newblock \emph{Proceedings of the IRE}, 49\penalty0 (1):\penalty0 8--30, 1961.
\newblock ISSN 00968390.
\newblock \doi{10.1109/JRPROC.1961.287775}.

\bibitem[Sutton(1984)]{sutton1984temporal}
Richard~S Sutton.
\newblock \emph{Temporal credit assignment in reinforcement learning}.
\newblock PhD thesis, University of Massachusetts, 1984.

\bibitem[Pignatelli et~al.(2024)Pignatelli, Ferret, Geist, Mesnard, van
  Hasselt, and Toni]{pignatelli2024a}
Eduardo Pignatelli, Johan Ferret, Matthieu Geist, Thomas Mesnard, Hado van
  Hasselt, and Laura Toni.
\newblock A survey of temporal credit assignment in deep reinforcement
  learning.
\newblock \emph{Transactions on Machine Learning Research}, 2024.
\newblock ISSN 2835-8856.
\newblock URL \url{https://openreview.net/forum?id=bNtr6SLgZf}.
\newblock Survey Certification.

\bibitem[Efroni et~al.(2021)Efroni, Merlis, and
  Mannor]{efroni2021reinforcement}
Yonathan Efroni, Nadav Merlis, and Shie Mannor.
\newblock Reinforcement learning with trajectory feedback.
\newblock In \emph{Proceedings of the AAAI Conference on Artificial
  Intelligence}, volume~35, pages 7288--7295, 2021.

\bibitem[Ng et~al.(1999)Ng, Harada, and Russell]{ng1999policy}
Andrew~Y Ng, Daishi Harada, and Stuart Russell.
\newblock Policy invariance under reward transformations: Theory and
  application to reward shaping.
\newblock In \emph{Icml}, volume~99, pages 278--287. Citeseer, 1999.

\bibitem[Gupta et~al.(2022)Gupta, Pacchiano, Zhai, Kakade, and
  Levine]{gupta2022unpacking}
Abhishek Gupta, Aldo Pacchiano, Yuexiang Zhai, Sham Kakade, and Sergey Levine.
\newblock Unpacking reward shaping: Understanding the benefits of reward
  engineering on sample complexity.
\newblock \emph{Advances in Neural Information Processing Systems},
  35:\penalty0 15281--15295, 2022.

\bibitem[Al-Emran(2015)]{al2015hierarchical}
Mostafa Al-Emran.
\newblock Hierarchical reinforcement learning: a survey.
\newblock \emph{International journal of computing and digital systems},
  4\penalty0 (02), 2015.

\bibitem[Sutton et~al.(1999)Sutton, Precup, and Singh]{sutton1999between}
Richard~S Sutton, Doina Precup, and Satinder Singh.
\newblock Between mdps and semi-mdps: A framework for temporal abstraction in
  reinforcement learning.
\newblock \emph{Artificial intelligence}, 112\penalty0 (1-2):\penalty0
  181--211, 1999.

\bibitem[Dalal et~al.(2024)Dalal, Chiruvolu, Chaplot, and
  Salakhutdinov]{dalal2024planseqlearn}
Murtaza Dalal, Tarun Chiruvolu, Devendra~Singh Chaplot, and Ruslan
  Salakhutdinov.
\newblock Plan-seq-learn: Language model guided {RL} for solving long horizon
  robotics tasks.
\newblock In \emph{The Twelfth International Conference on Learning
  Representations}, 2024.
\newblock URL \url{https://openreview.net/forum?id=hQVCCxQrYN}.

\bibitem[Klissarov et~al.(2023)Klissarov, D'Oro, Sodhani, Raileanu, Bacon,
  Vincent, Zhang, and Henaff]{klissarov2023motif}
Martin Klissarov, Pierluca D'Oro, Shagun Sodhani, Roberta Raileanu, Pierre-Luc
  Bacon, Pascal Vincent, Amy Zhang, and Mikael Henaff.
\newblock Motif: Intrinsic motivation from artificial intelligence feedback.
\newblock 2023.

\bibitem[Huang et~al.(2023)Huang, Xia, Shah, Driess, Zeng, Lu, Florence,
  Mordatch, Levine, Hausman, et~al.]{huang2023grounded}
Wenlong Huang, Fei Xia, Dhruv Shah, Danny Driess, Andy Zeng, Yao Lu, Pete
  Florence, Igor Mordatch, Sergey Levine, Karol Hausman, et~al.
\newblock Grounded decoding: Guiding text generation with grounded models for
  robot control.
\newblock \emph{arXiv preprint arXiv:2303.00855}, 2023.

\bibitem[Carta et~al.(2023)Carta, Romac, Wolf, Lamprier, Sigaud, and
  Oudeyer]{carta2023grounding}
Thomas Carta, Cl{\'e}ment Romac, Thomas Wolf, Sylvain Lamprier, Olivier Sigaud,
  and Pierre-Yves Oudeyer.
\newblock Grounding large language models in interactive environments with
  online reinforcement learning.
\newblock In \emph{International Conference on Machine Learning}, pages
  3676--3713. PMLR, 2023.

\bibitem[Jin et~al.(2023)Jin, Chen, Leeb, Gresele, Kamal, Lyu, Blin, Gonzalez,
  Kleiman-Weiner, Sachan, and Sch{\"{o}}lkopf]{jin2023cladder}
Zhijing Jin, Yuen Chen, Felix Leeb, Luigi Gresele, Ojasv Kamal, Zhiheng Lyu,
  Kevin Blin, Fernando Gonzalez, Max Kleiman-Weiner, Mrinmaya Sachan, and
  Bernhard Sch{\"{o}}lkopf.
\newblock {CL}adder: {A}ssessing causal reasoning in language models.
\newblock In \emph{NeurIPS}, 2023.
\newblock URL \url{https://openreview.net/forum?id=e2wtjx0Yqu}.

\bibitem[K{\i}c{\i}man et~al.(2023)K{\i}c{\i}man, Ness, Sharma, and
  Tan]{kiciman2023causal}
Emre K{\i}c{\i}man, Robert Ness, Amit Sharma, and Chenhao Tan.
\newblock Causal reasoning and large language models: Opening a new frontier
  for causality.
\newblock \emph{arXiv preprint arXiv:2305.00050}, 2023.

\bibitem[Samvelyan et~al.(2021)Samvelyan, Kirk, Kurin, Parker-Holder, Jiang,
  Hambro, Petroni, Kuttler, Grefenstette, and
  Rockt{\"a}schel]{samvelyan2021minihack}
Mikayel Samvelyan, Robert Kirk, Vitaly Kurin, Jack Parker-Holder, Minqi Jiang,
  Eric Hambro, Fabio Petroni, Heinrich Kuttler, Edward Grefenstette, and Tim
  Rockt{\"a}schel.
\newblock Minihack the planet: A sandbox for open-ended reinforcement learning
  research.
\newblock In \emph{Thirty-fifth Conference on Neural Information Processing
  Systems Datasets and Benchmarks Track (Round 1)}, 2021.
\newblock URL \url{https://openreview.net/forum?id=skFwlyefkWJ}.

\bibitem[Paischer et~al.(2022)Paischer, Adler, Patil, Bitto-Nemling,
  Holzleitner, Lehner, Eghbal-Zadeh, and Hochreiter]{paischer2022history}
Fabian Paischer, Thomas Adler, Vihang Patil, Angela Bitto-Nemling, Markus
  Holzleitner, Sebastian Lehner, Hamid Eghbal-Zadeh, and Sepp Hochreiter.
\newblock History compression via language models in reinforcement learning.
\newblock In \emph{International Conference on Machine Learning}, pages
  17156--17185. PMLR, 2022.

\bibitem[Paischer et~al.(2024)Paischer, Adler, Hofmarcher, and
  Hochreiter]{paischer2024semantic}
Fabian Paischer, Thomas Adler, Markus Hofmarcher, and Sepp Hochreiter.
\newblock Semantic helm: A human-readable memory for reinforcement learning.
\newblock \emph{Advances in Neural Information Processing Systems}, 36, 2024.

\bibitem[Yao et~al.(2020)Yao, Rao, Hausknecht, and Narasimhan]{yao2020keep}
Shunyu Yao, Rohan Rao, Matthew Hausknecht, and Karthik Narasimhan.
\newblock Keep calm and explore: Language models for action generation in
  text-based games.
\newblock \emph{arXiv preprint arXiv:2010.02903}, 2020.

\bibitem[Du et~al.(2023)Du, Watkins, Wang, Colas, Darrell, Abbeel, Gupta, and
  Andreas]{du2023guiding}
Yuqing Du, Olivia Watkins, Zihan Wang, C{\'e}dric Colas, Trevor Darrell, Pieter
  Abbeel, Abhishek Gupta, and Jacob Andreas.
\newblock Guiding pretraining in reinforcement learning with large language
  models.
\newblock In \emph{International Conference on Machine Learning}, pages
  8657--8677. PMLR, 2023.

\bibitem[Huang et~al.(2022)Huang, Abbeel, Pathak, and
  Mordatch]{huang2022language}
Wenlong Huang, Pieter Abbeel, Deepak Pathak, and Igor Mordatch.
\newblock Language models as zero-shot planners: Extracting actionable
  knowledge for embodied agents.
\newblock In \emph{International Conference on Machine Learning}, pages
  9118--9147. PMLR, 2022.

\bibitem[Wang et~al.(2023)Wang, Cai, Chen, Liu, Ma, and
  Liang]{wang2023describe}
Zihao Wang, Shaofei Cai, Guanzhou Chen, Anji Liu, Xiaojian Ma, and Yitao Liang.
\newblock Describe, explain, plan and select: Interactive planning with large
  language models enables open-world multi-task agents.
\newblock \emph{arXiv preprint arXiv:2302.01560}, 2023.

\bibitem[Singh et~al.(2023)Singh, Blukis, Mousavian, Goyal, Xu, Tremblay, Fox,
  Thomason, and Garg]{singh2023progprompt}
Ishika Singh, Valts Blukis, Arsalan Mousavian, Ankit Goyal, Danfei Xu, Jonathan
  Tremblay, Dieter Fox, Jesse Thomason, and Animesh Garg.
\newblock Progprompt: Generating situated robot task plans using large language
  models.
\newblock In \emph{2023 IEEE International Conference on Robotics and
  Automation (ICRA)}, pages 11523--11530. IEEE, 2023.

\bibitem[Brohan et~al.(2023)Brohan, Chebotar, Finn, Hausman, Herzog, Ho, Ibarz,
  Irpan, Jang, Julian, et~al.]{brohan2023can}
Anthony Brohan, Yevgen Chebotar, Chelsea Finn, Karol Hausman, Alexander Herzog,
  Daniel Ho, Julian Ibarz, Alex Irpan, Eric Jang, Ryan Julian, et~al.
\newblock Do as i can, not as i say: Grounding language in robotic affordances.
\newblock In \emph{Conference on robot learning}, pages 287--318. PMLR, 2023.

\bibitem[Dasgupta et~al.(2023)Dasgupta, Kaeser-Chen, Marino, Ahuja, Babayan,
  Hill, and Fergus]{dasgupta2023collaborating}
Ishita Dasgupta, Christine Kaeser-Chen, Kenneth Marino, Arun Ahuja, Sheila
  Babayan, Felix Hill, and Rob Fergus.
\newblock Collaborating with language models for embodied reasoning.
\newblock \emph{arXiv preprint arXiv:2302.00763}, 2023.

\bibitem[Shah et~al.(2023)Shah, Osi{\'n}ski, Levine, et~al.]{shah2023lm}
Dhruv Shah, B{\l}a{\.z}ej Osi{\'n}ski, Sergey Levine, et~al.
\newblock Lm-nav: Robotic navigation with large pre-trained models of language,
  vision, and action.
\newblock In \emph{Conference on robot learning}, pages 492--504. PMLR, 2023.

\bibitem[Zhong et~al.(2020)Zhong, Rockt{\"a}schel, and
  Grefenstette]{zhong2020rtfm}
Victor Zhong, Tim Rockt{\"a}schel, and Edward Grefenstette.
\newblock Rtfm: Generalising to new environment dynamics via reading.
\newblock In \emph{ICLR}, pages 1--17. ICLR, 2020.

\bibitem[Zhong et~al.(2022)Zhong, Mu, Zettlemoyer, Grefenstette, and
  Rockt{\"a}schel]{zhong2022improving}
Victor Zhong, Jesse Mu, Luke Zettlemoyer, Edward Grefenstette, and Tim
  Rockt{\"a}schel.
\newblock Improving policy learning via language dynamics distillation.
\newblock \emph{Advances in Neural Information Processing Systems},
  35:\penalty0 12504--12515, 2022.

\bibitem[Liang et~al.(2022)Liang, Huang, Xia, Xu, Hausman, Ichter, Florence,
  and Zeng]{liang2022code}
Jacky Liang, Wenlong Huang, Fei Xia, Peng Xu, Karol Hausman, Brian Ichter, Pete
  Florence, and Andy Zeng.
\newblock Code as policies: Language model programs for embodied control, 2022.
\newblock \emph{URL https://arxiv. org/abs/2209.07753}, 3, 2022.

\bibitem[Carta et~al.(2022)Carta, Oudeyer, Sigaud, and
  Lamprier]{carta2022eager}
Thomas Carta, Pierre-Yves Oudeyer, Olivier Sigaud, and Sylvain Lamprier.
\newblock Eager: Asking and answering questions for automatic reward shaping in
  language-guided rl.
\newblock \emph{Advances in Neural Information Processing Systems},
  35:\penalty0 12478--12490, 2022.

\bibitem[Goyal et~al.(2019)Goyal, Niekum, and Mooney]{goyal2019using}
Prasoon Goyal, Scott Niekum, and Raymond~J Mooney.
\newblock Using natural language for reward shaping in reinforcement learning.
\newblock \emph{arXiv preprint arXiv:1903.02020}, 2019.

\bibitem[Kwon et~al.(2023)Kwon, Xie, Bullard, and Sadigh]{kwon2023reward}
Minae Kwon, Sang~Michael Xie, Kalesha Bullard, and Dorsa Sadigh.
\newblock Reward design with language models, 2023.

\bibitem[Wu et~al.(2024)Wu, Fan, Liang, Azaria, Li, and Mitchell]{wu2024read}
Yue Wu, Yewen Fan, Paul~Pu Liang, Amos Azaria, Yuanzhi Li, and Tom~M Mitchell.
\newblock Read and reap the rewards: Learning to play atari with the help of
  instruction manuals.
\newblock \emph{Advances in Neural Information Processing Systems}, 36, 2024.

\bibitem[Bellemare et~al.(2013)Bellemare, Naddaf, Veness, and
  Bowling]{bellemare2013arcade}
Marc~G Bellemare, Yavar Naddaf, Joel Veness, and Michael Bowling.
\newblock The arcade learning environment: An evaluation platform for general
  agents.
\newblock \emph{Journal of Artificial Intelligence Research}, 47:\penalty0
  253--279, 2013.

\bibitem[K{\"u}ttler et~al.(2020)K{\"u}ttler, Nardelli, Miller, Raileanu,
  Selvatici, Grefenstette, and Rockt{\"a}schel]{kuttler2020nethack}
Heinrich K{\"u}ttler, Nantas Nardelli, Alexander Miller, Roberta Raileanu,
  Marco Selvatici, Edward Grefenstette, and Tim Rockt{\"a}schel.
\newblock The nethack learning environment.
\newblock \emph{Advances in Neural Information Processing Systems},
  33:\penalty0 7671--7684, 2020.

\bibitem[Hung et~al.(2019)Hung, Lillicrap, Abramson, Wu, Mirza, Carnevale,
  Ahuja, and Wayne]{hung2019optimizing}
Chia-Chun Hung, Timothy Lillicrap, Josh Abramson, Yan Wu, Mehdi Mirza, Federico
  Carnevale, Arun Ahuja, and Greg Wayne.
\newblock Optimizing agent behavior over long time scales by transporting
  value.
\newblock \emph{Nature Communications}, 10\penalty0 (1):\penalty0 5223, 2019.
\newblock ISSN 2041-1723.
\newblock \doi{10.1038/s41467-019-13073-w}.
\newblock URL \url{https://doi.org/10.1038/s41467-019-13073-w}.

\bibitem[Mesnard et~al.(2021)Mesnard, Weber, Viola, Thakoor, Saade,
  Harutyunyan, Dabney, Stepleton, Heess, Guez,
  et~al.]{mesnard2021counterfactual}
Thomas Mesnard, Theophane Weber, Fabio Viola, Shantanu Thakoor, Alaa Saade,
  Anna Harutyunyan, Will Dabney, Thomas~S Stepleton, Nicolas Heess, Arthur
  Guez, et~al.
\newblock Counterfactual credit assignment in model-free reinforcement
  learning.
\newblock In \emph{International Conference on Machine Learning}, pages
  7654--7664. Proceedings of Machine Learning Research, 2021.

\bibitem[Jiang et~al.(2019)Jiang, Gu, Murphy, and Finn]{jiang2019language}
Yiding Jiang, Shixiang~Shane Gu, Kevin~P Murphy, and Chelsea Finn.
\newblock Language as an abstraction for hierarchical deep reinforcement
  learning.
\newblock \emph{Advances in Neural Information Processing Systems}, 32, 2019.

\bibitem[Jacob et~al.(2021)Jacob, Lewis, and Andreas]{jacob2021multitasking}
Athul~Paul Jacob, Mike Lewis, and Jacob Andreas.
\newblock Multitasking inhibits semantic drift.
\newblock \emph{arXiv preprint arXiv:2104.07219}, 2021.

\bibitem[Sharma et~al.(2021)Sharma, Torralba, and Andreas]{sharma2021skill}
Pratyusha Sharma, Antonio Torralba, and Jacob Andreas.
\newblock Skill induction and planning with latent language.
\newblock \emph{arXiv preprint arXiv:2110.01517}, 2021.

\bibitem[Mu et~al.(2022)Mu, Zhong, Raileanu, Jiang, Goodman, Rockt{\"a}schel,
  and Grefenstette]{mu2022improving}
Jesse Mu, Victor Zhong, Roberta Raileanu, Minqi Jiang, Noah Goodman, Tim
  Rockt{\"a}schel, and Edward Grefenstette.
\newblock Improving intrinsic exploration with language abstractions.
\newblock \emph{Advances in Neural Information Processing Systems},
  35:\penalty0 33947--33960, 2022.

\bibitem[Mesnard et~al.(2023)Mesnard, Chen, Saade, Tang, Rowland, Weber, Lyle,
  Gruslys, Valko, Dabney, Ostrovski, Moulines, and Munos]{mesnard2023quantile}
Thomas Mesnard, Wenqi Chen, Alaa Saade, Yunhao Tang, Mark Rowland, Theophane
  Weber, Clare Lyle, Audrunas Gruslys, Michal Valko, Will Dabney, Georg
  Ostrovski, Eric Moulines, and Remi Munos.
\newblock Quantile credit assignment.
\newblock In Andreas Krause, Emma Brunskill, Kyunghyun Cho, Barbara Engelhardt,
  Sivan Sabato, and Jonathan Scarlett, editors, \emph{Proceedings of the 40th
  International Conference on Machine Learning}, volume 202 of
  \emph{Proceedings of Machine Learning Research}, pages 24517--24531. PMLR,
  23--29 Jul 2023.
\newblock URL \url{https://proceedings.mlr.press/v202/mesnard23a.html}.

\bibitem[Chevalier-Boisvert et~al.(2018)Chevalier-Boisvert, Willems, and
  Pal]{chevalier2018gym_minigrid}
Maxime Chevalier-Boisvert, Lucas Willems, and Suman Pal.
\newblock Minimalistic gridworld environment for openai gym.
\newblock \url{https://github.com/maximecb/gym-minigrid}, 2018.

\bibitem[Dean et~al.(2012)Dean, Corrado, Monga, Chen, Devin, Mao, Ranzato,
  Senior, Tucker, Yang, et~al.]{dean2012large}
Jeffrey Dean, Greg Corrado, Rajat Monga, Kai Chen, Matthieu Devin, Mark Mao,
  Marc'aurelio Ranzato, Andrew Senior, Paul Tucker, Ke~Yang, et~al.
\newblock Large scale distributed deep networks.
\newblock \emph{Advances in neural information processing systems}, 25, 2012.

\bibitem[Dettmers et~al.(2023)Dettmers, Pagnoni, Holtzman, and
  Zettlemoyer]{dettmers2023qlora}
Tim Dettmers, Artidoro Pagnoni, Ari Holtzman, and Luke Zettlemoyer.
\newblock Qlora: Efficient finetuning of quantized llms. arxiv 2023.
\newblock \emph{arXiv preprint arXiv:2305.14314}, 2023.

\bibitem[Jeurissen et~al.(2024)Jeurissen, Perez-Liebana, Gow, Cakmak, and
  Kwan]{jeurissen2024playing}
Dominik Jeurissen, Diego Perez-Liebana, Jeremy Gow, Duygu Cakmak, and James
  Kwan.
\newblock Playing nethack with llms: Potential \& limitations as zero-shot
  agents.
\newblock \emph{arXiv preprint arXiv:2403.00690}, 2024.

\bibitem[Hafner(2021)]{hafner2021crafter}
Danijar Hafner.
\newblock Benchmarking the spectrum of agent capabilities.
\newblock \emph{arXiv preprint arXiv:2109.06780}, 2021.

\bibitem[Matthews et~al.(2024)Matthews, Beukman, Ellis, Samvelyan, Jackson,
  Coward, and Foerster]{matthews2024craftax}
Michael Matthews, Michael Beukman, Benjamin Ellis, Mikayel Samvelyan, Matthew
  Jackson, Samuel Coward, and Jakob Foerster.
\newblock Craftax: A lightning-fast benchmark for open-ended reinforcement
  learning.
\newblock \emph{arXiv preprint arXiv:2402.16801}, 2024.

\bibitem[Baumli et~al.(2023)Baumli, Baveja, Behbahani, Chan, Comanici,
  Flennerhag, Gazeau, Holsheimer, Horgan, Laskin, et~al.]{baumli2023vision}
Kate Baumli, Satinder Baveja, Feryal Behbahani, Harris Chan, Gheorghe Comanici,
  Sebastian Flennerhag, Maxime Gazeau, Kristian Holsheimer, Dan Horgan, Michael
  Laskin, et~al.
\newblock Vision-language models as a source of rewards.
\newblock \emph{arXiv preprint arXiv:2312.09187}, 2023.

\end{thebibliography}

\appendix
\section{Prompting}
\label{app:prompts}
To develop an intuition of the \gls{llm}'s task, we show examples of prompts for each configuration used in the experiments of Section~\ref{sec:res:preset}.
In particular, we show examples of prompts for both option termination verification and option discovery, and both game screen and cropped observations.
Finally, we present prompts where two options terminated (a subgoal is achieved): a key pick-up and a door unlock.

\subsection{Cropped vs gamescreen observations}
\begin{prompt}[h]
\begin{llmbox}[frametitle=Prompt with cropped observations]
\begin{verbatim}
The environment is MiniHack.

I will present you with a short extract of a gameplay. At each timestep, symbols represent the following items:
- "." represents a floor tile.
- "|" can represent either a wall, a vertical wall, an open door.
- "-" can represent either the bottom left corner (of a room), bottom right corner (of a room), wall, horizontal wall, wall, top left corner (of a room), op right corner (of a room).
- "+" represents a closed door. Doors can be locked, and require a key to open.
- "(" represents a useful item (pick-axe, key, lamp...)
- "<" represents a ladder or staircase up.
- ">" represents a ladder or staircase down.

The task of the agent is to win the game.

First, based on your knowledge of NetHack, break down the task of the agent into subgoals. 
Then, consider the following game transition, which might or might not contain these subgoals.
Determine if any of the subgoals is achieved at Time: 1 or not.


Report your response in a dictionary containing the name of the subgoals as keys and booleans as value. For example:
```python
{
    <name of goal>: <bool>,
} 

Observation Sequence:

<gameplay>
Time: 0
Current message: Never mind.

                 
      - - - -    
      | . . |    
      | . . |    
      - @ < |    
  . . . . . |    
| . ( . . . |    
- - - - - - -    
                 
Time: 1
Current message: 

                 
                 
      - - - -    
      | . . |    
      | @ . |    
      - . < |    
  . . . . . |    
| . ( . . . |    
- - - - - - -    

</gameplay>

I will not consider anything that is not in the dictionary.
You have only one shot at this, and you cannot ask for clarifications.
\end{verbatim}
\end{llmbox}
\caption{Example of a prompt where transitions have cropped observations.}
\label{app:prompt:cropped}
\end{prompt}

\begin{prompt}[H]
\begin{llmbox}[frametitle=Prompt with gamescreen observations]
\begin{verbatim}
The environment is MiniHack.

I will present you with a short extract of a gameplay. At each timestep, symbols represent the following items:
- "." represents a floor tile.
- "|" can represent either a wall, a vertical wall, an open door.
- "-" can represent either the bottom left corner (of a room), bottom right corner (of a room), wall, horizontal wall, wall, top left corner (of a room), op right corner (of a room).
- "+" represents a closed door. Doors can be locked, and require a key to open.
- "(" represents a useful item (pick-axe, key, lamp...)
- "<" represents a ladder or staircase up.
- ">" represents a ladder or staircase down.

The task of the agent is to win the game.

First, based on your knowledge of NetHack, break down the task of the agent into subgoals. 
Then, consider the following game transition, which might or might not contain these subgoals.
Determine if any of the subgoals is achieved at Time: 1 or not.


Report your response in a dictionary containing the name of the subgoals as keys and booleans as value. For example:
```python
{
    <name of goal>: <bool>,
} 

Observation Sequence:

<gameplay>
Time: 0

Never mind.                                                                     
                                                                                                                                                               
                                                                                                                                                               
                                                                                                                                                               
                                                                                                                                                               
                                                                                                                                                               
                                                                                                                                                               
                                                                                                                                                               
                                                                                                                                                               
                                                                              - - - -                                                                          
                                                                              | . . |                                                                          
                                                                              | . . |                                                                          
                                                                              - @ < |                                                                          
                                                                          . . . . . |                                                                          
                                                                        | . ( . . . |                                                                          
                                                                        - - - - - - -                                                                          
                                                                                                                                                               
                                                                                                                                                               
                                                                                                                                                               
                                                                                                                                                               
                                                                                                                                                               
                                                                                
Agent the Footpad              St:14 Dx:17 Co:17 In:9 Wi:11 Ch:7 Chaotic S:0    
Dlvl:1 $:0 HP:12(12) Pw:2(2) AC:7 Xp:1/0                                        
Time: 1

                                                                                
                                                                                                                                                               
                                                                                                                                                               
                                                                                                                                                               
                                                                                                                                                               
                                                                                                                                                               
                                                                                                                                                               
                                                                                                                                                               
                                                                                                                                                               
                                                                              - - - -                                                                          
                                                                              | . . |                                                                          
                                                                              | @ . |                                                                          
                                                                              - . < |                                                                          
                                                                          . . . . . |                                                                          
                                                                        | . ( . . . |                                                                          
                                                                        - - - - - - -                                                                          
                                                                                                                                                               
                                                                                                                                                               
                                                                                                                                                               
                                                                                                                                                               
                                                                                                                                                               
                                                                                
Agent the Footpad              St:14 Dx:17 Co:17 In:9 Wi:11 Ch:7 Chaotic S:0    
Dlvl:1 $:0 HP:12(12) Pw:2(2) AC:7 Xp:1/0                                        

</gameplay>

I will not consider anything that is not in the dictionary.
You have only one shot at this, and you cannot ask for clarifications.
\end{verbatim}
\end{llmbox}
\caption{Example of a prompt where transitions have game screen observations.}
\label{app:prompt:gamescreen}
\end{prompt}
\FloatBarrier

\subsection{Option termination vs option discovery}
\begin{prompt}[h]
\begin{llmbox}[frametitle=Prompt with cropped observations]
\begin{verbatim}
The environment is MiniHack.

I will present you with a short extract of a gameplay. At each timestep, symbols represent the following items:
- "." represents a floor tile.
- "|" can represent either a wall, a vertical wall, an open door.
- "-" can represent either the bottom left corner (of a room), bottom right corner (of a room), wall, horizontal wall, wall, top left corner (of a room), op right corner (of a room).
- "+" represents a closed door. Doors can be locked, and require a key to open.
- "(" represents a useful item (pick-axe, key, lamp...)
- "<" represents a ladder or staircase up.
- ">" represents a ladder or staircase down.

The task of the agent is to win the game.

Consider the following subgoals:
```python
subgoals = {
    "pick up the key": None,
    "open the door": None,
} 
``` 
Then, consider the following game transition, which might or might not contain these subgoals.
Determine if any of the subgoals is achieved at Time: 1 or not.


Report your response in a dictionary containing the name of the subgoals as keys and booleans as value. For example:
```python
{
    <name of goal>: <bool>,
} 

Observation Sequence:

<gameplay>
Time: 0
Current message: Never mind.

                 
      - - - -    
      | . . |    
      | . . |    
      - @ < |    
  . . . . . |    
| . ( . . . |    
- - - - - - -    
                 
Time: 1
Current message: 

                 
                 
      - - - -    
      | . . |    
      | @ . |    
      - . < |    
  . . . . . |    
| . ( . . . |    
- - - - - - -    

</gameplay>

I will not consider anything that is not in the dictionary.
You have only one shot at this, and you cannot ask for clarifications.
\end{verbatim}
\end{llmbox}
\caption{Example of a prompt for instruction verification. Here, goals are provided externally from a human.}
\label{app:prompt:termination}
\end{prompt}

\begin{prompt}[h]
\begin{llmbox}[frametitle=Prompt with cropped observations]
\begin{verbatim}
The environment is MiniHack.

I will present you with a short extract of a gameplay. At each timestep, symbols represent the following items:
- "." represents a floor tile.
- "|" can represent either a wall, a vertical wall, an open door.
- "-" can represent either the bottom left corner (of a room), bottom right corner (of a room), wall, horizontal wall, wall, top left corner (of a room), op right corner (of a room).
- "+" represents a closed door. Doors can be locked, and require a key to open.
- "(" represents a useful item (pick-axe, key, lamp...)
- "<" represents a ladder or staircase up.
- ">" represents a ladder or staircase down.

The task of the agent is to win the game.

First, based on your knowledge of NetHack, break down the task of the agent into subgoals. 
Then, consider the following game transition, which might or might not contain these subgoals.
Determine if any of the subgoals is achieved at Time: 1 or not.


Report your response in a dictionary containing the name of the subgoals as keys and booleans as value. For example:
```python
{
    <name of goal>: <bool>,
} 

Observation Sequence:

<gameplay>
Time: 0
Current message: Never mind.

                 
      - - - -    
      | . . |    
      | . . |    
      - @ < |    
  . . . . . |    
| . ( . . . |    
- - - - - - -    
                 
Time: 1
Current message: 

                 
                 
      - - - -    
      | . . |    
      | @ . |    
      - . < |    
  . . . . . |    
| . ( . . . |    
- - - - - - -    

</gameplay>

I will not consider anything that is not in the dictionary.
You have only one shot at this, and you cannot ask for clarifications.
\end{verbatim}
\end{llmbox}
\caption{Example of a prompt for options discovery and termination verification.}
\label{app:prompt:discovery}
\end{prompt}
\FloatBarrier

\subsection{Examples of different subgoals}
\begin{prompt}[h]
\begin{llmbox}[frametitle=Examples of key pickup]
\begin{verbatim}
The environment is MiniHack.

I will present you with a short extract of a gameplay. At each timestep, symbols represent the following items:
- "." represents a floor tile.
- "|" can represent either a wall, a vertical wall, an open door.
- "-" can represent either the bottom left corner (of a room), bottom right corner (of a room), wall, horizontal wall, wall, top left corner (of a room), op right corner (of a room).
- "+" represents a closed door. Doors can be locked, and require a key to open.
- "(" represents a useful item (pick-axe, key, lamp...)
- "<" represents a ladder or staircase up.
- ">" represents a ladder or staircase down.

The task of the agent is to win the game.

First, based on your knowledge of NetHack, break down the task of the agent into subgoals. 
Then, consider the following game transition, which might or might not contain these subgoals.
Determine if any of the subgoals is achieved at Time: 1 or not.


Report your response in a dictionary containing the name of the subgoals as keys and booleans as value. For example:
```python
{
    <name of goal>: <bool>,
} 

Observation Sequence:

<gameplay>
Time: 0
Current message: It's a wall.

          | . . |
          | . . |
    - - + - . < |
    | . . . . . |
    | . @ . . . |
    - - - - - - -
                 
                 
                 
Time: 1
Current message: g - a key named The Master Key of Thievery.

          | . . |
          | . . |
    - - + - . < |
    | . . . . . |
    | . @ . . . |
    - - - - - - -
                 
                 
                 

</gameplay>

I will not consider anything that is not in the dictionary.
You have only one shot at this, and you cannot ask for clarifications.
\end{verbatim}
\end{llmbox}
\caption{Example of a prompt where the transition shows a successful key pickup subgoal.}
\label{app:prompt:key-pickup}
\end{prompt}

\begin{prompt}[h]
\begin{llmbox}[frametitle=Examples of door unlock]
\begin{verbatim}

The environment is MiniHack.

I will present you with a short extract of a gameplay. At each timestep, symbols represent the following items:
- "." represents a floor tile.
- "|" can represent either a wall, a vertical wall, an open door.
- "-" can represent either the bottom left corner (of a room), bottom right corner (of a room), wall, horizontal wall, wall, top left corner (of a room), op right corner (of a room).
- "+" represents a closed door. Doors can be locked, and require a key to open.
- "(" represents a useful item (pick-axe, key, lamp...)
- "<" represents a ladder or staircase up.
- ">" represents a ladder or staircase down.

The task of the agent is to win the game.

First, based on your knowledge of NetHack, break down the task of the agent into subgoals. 
Then, consider the following game transition, which might or might not contain these subgoals.
Determine if any of the subgoals is achieved at Time: 1 or not.


Report your response in a dictionary containing the name of the subgoals as keys and booleans as value. For example:
```python
{
    <name of goal>: <bool>,
} 

Observation Sequence:

<gameplay>
Time: 0
Current message: 

        - - - -  
        | . . |  
        | . . |  
  - - + - . < |  
  | . . @ . . |  
  | . . . . . |  
  - - - - - - -  
                 
                 
Time: 1
Current message: Never mind.

  - -   - - - -  
  | >   | . . |  
    . . | . . |  
  - - | - . < |  
  | . . @ . . |  
  | . . . . . |  
  - - - - - - -  
                 
                 

</gameplay>

I will not consider anything that is not in the dictionary.
You have only one shot at this, and you cannot ask for clarifications.
\end{verbatim}
\end{llmbox}
\caption{Example of a prompt where the transition shows a successful door-unlock subgoal.}
\label{app:prompt:door-unlock}
\end{prompt}
\FloatBarrier

\section{Responses}
\label{app:responses-subgoals}
To give an intuition of how different models approach the task, here, we show a series of responses to Prompt~\ref{prompt:example}.
Notice that, to stimulate the model to reason about subgoals, we ask to consider its knowledge of Nethack, rather than MiniHack.

\begin{response}[h]
\begin{llmbox}[frametitle=c4ai-command-r-v01]
\begin{verbatim}
```python
{
    'find_item': False,
    'open_door': False,
    'collect_item': False,
    'move_upstairs': False,
    'move_downstairs': True,
    'kill_monster': False,
    'find_key': False,
    'open_lock': False,
    'light_lamp': False,
    'win_game': False
}
```  
\end{verbatim}
\end{llmbox}
\caption{Example response }
\end{response}

\begin{response}[h]
\begin{llmbox}[frametitle=gemma-1.1-7b-it]
\begin{verbatim}
## Subgoals:

- **Find a way out:** Locate and reach the exit of the map.
- **Gather useful items:** Collect any available items for future use.
- **Explore the environment:** Discover the layout of the map and potential hazards.


## Subgoal Achieved at Time: 1

```python
{
    "Find a way out": False,
    "Gather useful items": True,
    "Explore the environment": True,
}
```
\end{verbatim}
\end{llmbox}
\caption{Example response }
\end{response}

\begin{response}[h]
\begin{llmbox}[frametitle=Mistral-7B-Instruct-v0.2]
\begin{verbatim}
Based on my understanding of the given environment and the task, the subgoals for the agent are as follows:

1. Exploration: Reach all the unvisited parts of the map.
2. Collection: Gather all the useful items (pick-axe, key, lamp, etc.)
3. Door Opening: Open all the doors (locked or not) to progress through the map.

Now, let's analyze the given gameplay sequence and determine if any of the subgoals have been achieved at Time: 1.

```python
{
    "Exploration": False,
    "Collection": False,
    "Door Opening": False
}
```

At Time: 1, the agent has moved to a new location and has not interacted with any items or doors. Therefore, none of the subgoals have been achieved at this point.
\end{verbatim}
\end{llmbox}
\caption{Example response }
\end{response}

\begin{response}[h]
\begin{llmbox}[frametitle=Mixtral-8x7B-Instruct-v0.1]
\begin{verbatim}
Based on my knowledge of NetHack, the task of the agent can be broken down into the following subgoals:

1. Movement: The agent needs to move through the game environment to reach the goal.
2. Collection: The agent needs to collect useful items that can help it in achieving the goal.
3. Door manipulation: The agent needs to open doors to explore new areas.
4. Staircase navigation: The agent needs to navigate staircases to move to different levels of the game.
5. Key collection: The agent needs to collect keys to open locked doors.
6. Winning the game: The agent needs to reach the final goal and win the game.

Based on the given game transition, the subgoals achieved at Time: 1 are:

```python
{
    "Movement": True,
    "Collection": False,
    "Door manipulation": False,
    "Staircase navigation": False,
    "Key collection": False,
    "Winning the game": False,
}
```

Explanation:

* Movement: The agent moved from the bottom left corner of the room to the bottom right corner of the room, so this subgoal is achieved.
* Collection: The agent did not collect any item in this transition, so this subgoal is not achieved.
* Door manipulation: The agent did not encounter any doors in this transition, so this subgoal is not achieved.
* Staircase navigation: The agent did not encounter any staircases in this transition, so this subgoal is not achieved.
* Key collection: The agent did not collect any keys in this transition, so this subgoal is not achieved.
* Winning the game: The agent did not reach the final goal in this transition, so this subgoal is not achieved.
\end{verbatim}
\end{llmbox}
\caption{Example response }
\end{response}

\begin{response}[h]
\begin{llmbox}[frametitle=Llama-3-8B-Instruct]
\begin{verbatim}
Based on my knowledge of NetHack, I break down the task of the agent into the following subgoals:

* Find a useful item (pick-axe, key, lamp...)
* Find a ladder or staircase up
* Find a ladder or staircase down
* Open a door (if it's closed)
* Unlock a door (if it's locked)
* Reach the goal (win the game)

Now, considering the given game transition, I determine if any of the subgoals has been achieved at Time: 1 or not:

{
    'Find a useful item': False,
    'Find a ladder or staircase up': False,
    'Find a ladder or staircase down': True,
    'Open a door (if it\'s closed)': False,
    'Unlock a door (if it\'s locked)': False,
    'Reach the goal': False
}

Note that the agent has moved north and then east, but no new items or doors have been discovered, and no doors have been opened or unlocked. However, the agent is now closer to the ladder or staircase down, which is a step towards reaching the goal.
\end{verbatim}
\end{llmbox}
\caption{Example response }
\end{response}

\begin{response}[h]
\begin{llmbox}[frametitle=Llama-3-70B-Instruct]
\begin{verbatim}
Based on my knowledge of NetHack, I break down the task of the agent into the following subgoals:

* Reach the staircase/ladder: The agent needs to find a way to move up or down the levels to progress in the game.
* Collect useful items: The agent needs to collect items such as pick-axes, keys, and lamps to overcome obstacles and access new areas.
* Unlock doors: The agent needs to find keys to unlock closed doors and access new areas.
* Avoid obstacles: The agent needs to avoid walls, locked doors, and other obstacles that can hinder its progress.
* Reach the goal: The agent needs to reach the final goal, which is not specified in this extract but is typically the treasure or the exit.

Now, considering the given game transition, I determine if any of the subgoals have been achieved at Time: 1 or not:

{
    'Reach the staircase/ladder': False,
    'Collect useful items': False,
    'Unlock doors': False,
    'Avoid obstacles': True,
    'Reach the goal': False,
}

The agent has avoided obstacles by moving north and not hitting the wall, but none of the other subgoals have been achieved at Time: 1.    
\end{verbatim}
\end{llmbox}
\caption{Example prompt for option discovery, where goals are not provided externally.}
\end{response}
\FloatBarrier

\section{Scale}
In this section we show how the F1 score varies according to the number of parameters of the \gls{llm}.
For a fair analysis, we only compare models of the same family, where multiple sizes of the model are available.
Namely, we consider: \textit{Meta-Llama-3-8B-Instruct} and \textit{Meta-Llama-3-70B-Instruct}; \textit{gemma-1.1-2b-it} and \textit{gemma-1.1-7b-it}; \textit{Mistral-7B-Instruct-v0.2} and \textit{Mixtral-8x7B-Instruct-v0.1}.
Since \textit{c4ai-command-r-plus} does not fit an Nvidia A100 80Gb, we exclude \textit{c4ai-command-r-v01} from the comparison.
\begin{figure}[h]
    \centering
    \begin{subfigure}[b]{0.45\textwidth}
        \centering
        \includegraphics[width=\linewidth]{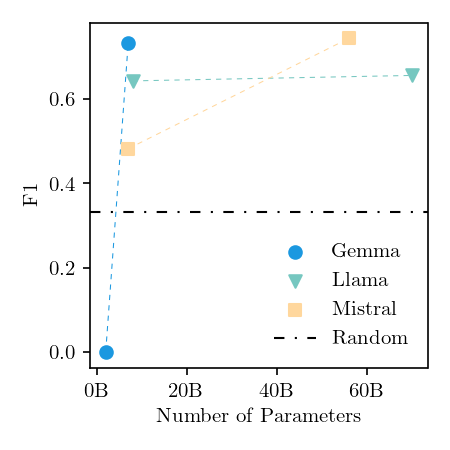}
        \caption{}
        \label{fig:dag:a}
    \end{subfigure}%
    \hfill
    \begin{subfigure}[b]{0.45\textwidth}
        \centering
        \includegraphics[width=\linewidth]{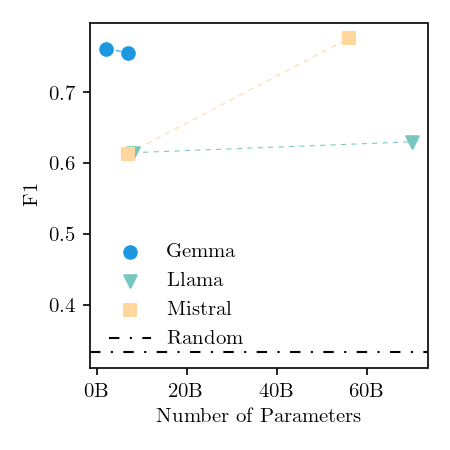}
        \caption{}
        \label{fig:dag:b}
    \end{subfigure}%
    \hfill
    \begin{subfigure}[b]{0.45\textwidth}
        \centering
        \includegraphics[width=\linewidth]{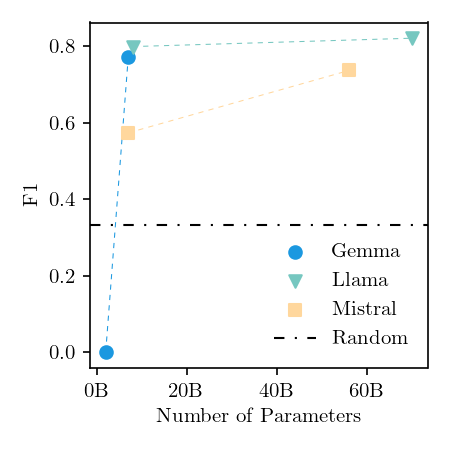}
        \caption{}
        \label{fig:dag:c}
    \end{subfigure}%
    \hfill
    \begin{subfigure}[b]{0.45\textwidth}
        \centering
        \includegraphics[width=\linewidth]{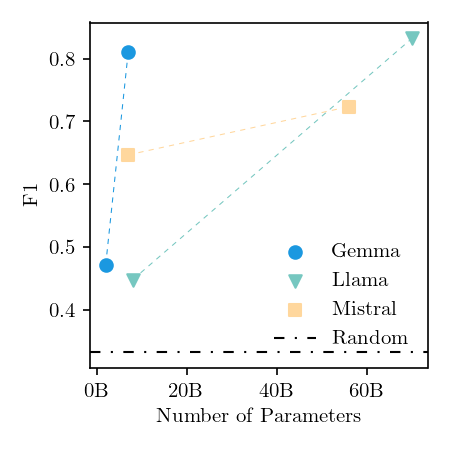}
        \caption{}
        \label{fig:enter-label}
    \end{subfigure}
\caption{F1 score as a function of the \gls{llm} size.}
\label{fig:dag}
\end{figure}
\FloatBarrier

\section{Ablations}
In this section, we include results of ablations, to understand which part of the prompt affect the performance the most.
We present two ablations:
\begin{enumerate}[label=\textit{(\roman*)}]
    \item With and without a token separator to isolate each cell in the grid observation.
    \item With and without including actions in the transition.
\end{enumerate}

\subsection{Tokenisation}
\label{app:ablations:tokens}
\begin{prompt}[h]
\begin{llmbox}[frametitle=Examples of prompt with no token separator]
\begin{verbatim}
The environment is MiniHack.

I will present you with a short extract of a gameplay. At each timestep, symbols represent the following items:
- "." represents a floor tile.
- "|" can represent either a wall, a vertical wall, an open door.
- "-" can represent either the bottom left corner (of a room), bottom right corner (of a room), wall, horizontal wall, wall, top left corner (of a room), op right corner (of a room).
- "+" represents a closed door. Doors can be locked, and require a key to open.
- "(" represents a useful item (pick-axe, key, lamp...)
- "<" represents a ladder or staircase up.
- ">" represents a ladder or staircase down.

The task of the agent is to win the game.

First, based on your knowledge of NetHack, break down the task of the agent into subgoals. 
Then, consider the following game transition, which might or might not contain these subgoals.
Determine if any of the subgoals is achieved at Time: 1 or not.


Report your response in a dictionary containing the name of the subgoals as keys and booleans as value. For example:
```python
{
    <name of goal>: <bool>,
} 

Observation Sequence:

<gameplay>
Time: 0
Current message: Never mind.

         
   ----  
   |..|  
   |..|  
   -@<|  
 .....|  
|.(...|  
-------  
         
Time: 1
Current message: 

         
         
   ----  
   |..|  
   |@.|  
   -.<|  
 .....|  
|.(...|  
-------  

</gameplay>

I will not consider anything that is not in the dictionary.
You have only one shot at this, and you cannot ask for clarifications.
\end{verbatim}
\end{llmbox}
\caption{Example of a prompt where the transition shows a successful door-unlock subgoal.}
\label{fig:app:token-ablation}
\end{prompt}

\begin{table}[h]
\centering
\begin{adjustbox}{max width=1\textwidth,center}
\begin{tabular}{lrrrrrrrr}
Annotator & F1 $\uparrow$ & Accuracy $\uparrow$ & Precision $\uparrow$ & Recall $\uparrow$ & TP $\uparrow$ & TN $\uparrow$ & FP $\downarrow$ & FN $\downarrow$ \\
\midrule
Human & \textbf{1.00} & \textbf{1.00} & \textbf{1.00} & \textbf{1.00} & \textbf{171} & \textbf{85} & \textbf{0} & \textbf{0} \\
\midrule
Mixtral-8x7B-Instruct-v0.1 & \textbf{0.73} & \textbf{0.65} & 0.75 & \textbf{0.71} & \textbf{121} & 45 & 40 & \textbf{50} \\
c4ai-command-r-v01 & 0.66 & 0.63 & 0.87 & 0.53 & 90 & 72 & 13 & 81 \\
gemma-1.1-7b-it & 0.66 & 0.64 & 0.89 & 0.52 & 89 & 74 & 11 & 82 \\
Mistral-7B-Instruct-v0.2 & 0.58 & 0.60 & 0.97 & 0.41 & 70 & 83 & 2 & 101 \\
Meta-Llama-3-8B-Instruct & 0.54 & 0.56 & 0.90 & 0.39 & 66 & 78 & 7 & 105 \\
gemma-1.1-2b-it & 0.53 & 0.49 & 0.69 & 0.43 & 74 & 52 & 33 & 97 \\
Meta-Llama-3-70B-Instruct & 0.16 & 0.39 & \textbf{1.00} & 0.09 & 15 & \textbf{85} & \textbf{0} & 156 \\
c4ai-command-r-plus & 0.02 & 0.33 & 0.40 & 0.01 & 2 & 82 & 3 & 169 \\
\midrule
Random & 0.33 & 0.33 & 0.33 & 0.33 \\
\end{tabular}
\end{adjustbox}
\vspace{1em}
\caption{
Performance of \gls{llm} annotations with \textbf{game screen} observations, subgoals \textbf{provided} in the prompt, and \textbf{no token separator}.
}
\vspace{-1em}
\label{tab:app:ablation-token-preset-gamescreen}
\end{table}

\begin{table}[h]
\centering
\begin{adjustbox}{max width=1\textwidth,center}
\begin{tabular}{lrrrrrrrr}
Annotator & F1 $\uparrow$ & Accuracy $\uparrow$ & Precision $\uparrow$ & Recall $\uparrow$ & TP $\uparrow$ & TN $\uparrow$ & FP $\downarrow$ & FN $\downarrow$ \\
\midrule
Human & \textbf{1.00} & \textbf{1.00} & \textbf{1.00} & \textbf{1.00} & \textbf{171} & \textbf{85} & \textbf{0} & \textbf{0} \\
\midrule
Mixtral-8x7B-Instruct-v0.1 & \textbf{0.76} & 0.68 & 0.77 & \textbf{0.75} & \textbf{128} & 47 & 38 & \textbf{43} \\
gemma-1.1-7b-it & \textbf{0.76} & 0.68 & 0.77 & \textbf{0.75} & \textbf{128} & 46 & 39 & \textbf{43} \\
c4ai-command-r-v01 & 0.74 & \textbf{0.69} & 0.83 & 0.67 & 115 & 62 & 23 & 56 \\
Meta-Llama-3-70B-Instruct & 0.71 & 0.68 & 0.89 & 0.60 & 102 & 72 & 13 & 69 \\
gemma-1.1-2b-it & 0.68 & 0.57 & 0.67 & 0.68 & 116 & 29 & 56 & 55 \\
c4ai-command-r-plus & 0.64 & 0.61 & 0.84 & 0.51 & 88 & 68 & 17 & 83 \\
Mistral-7B-Instruct-v0.2 & 0.59 & 0.61 & \textbf{0.99} & 0.42 & 72 & \textbf{84} & \textbf{1} & 99 \\
Meta-Llama-3-8B-Instruct & 0.49 & 0.52 & 0.87 & 0.34 & 58 & 76 & 9 & 113 \\
\midrule
Random & 0.33 & 0.33 & 0.33 & 0.33 \\
\end{tabular}
\end{adjustbox}
\vspace{1em}
\caption{
Performance of \gls{llm} annotations with \textbf{cropped} observations, subgoals \textbf{provided} in the prompt, and \textbf{no token separator}.
}
\vspace{-1em}
\label{tab:app:ablation-token-preset-cropped}
\end{table}

\begin{table}[h]
\centering
\begin{adjustbox}{max width=1\textwidth,center}
\begin{tabular}{lrrrrrrrr}
Annotator & F1 $\uparrow$ & Accuracy $\uparrow$ & Precision $\uparrow$ & Recall $\uparrow$ & TP $\uparrow$ & TN $\uparrow$ & FP $\downarrow$ & FN $\downarrow$ \\
\midrule
Human & \textbf{1.00} & \textbf{1.00} & \textbf{1.00} & \textbf{1.00} & \textbf{171} & \textbf{85} & \textbf{0} & \textbf{0} \\
\midrule
Meta-Llama-3-70B-Instruct & \textbf{0.81} & 0.71 & 0.72 & \textbf{0.92} & \textbf{157} & 25 & 60 & \textbf{14} \\
c4ai-command-r-plus & 0.79 & \textbf{0.74} & \textbf{0.86} & 0.73 & 125 & 64 & 21 & 46 \\
gemma-1.1-7b-it & 0.75 & 0.63 & 0.68 & 0.84 & 144 & 17 & 68 & 27 \\
Mixtral-8x7B-Instruct-v0.1 & 0.72 & 0.61 & 0.69 & 0.76 & 130 & 27 & 58 & 41 \\
c4ai-command-r-v01 & 0.63 & 0.58 & 0.76 & 0.54 & 92 & 56 & 29 & 79 \\
Meta-Llama-3-8B-Instruct & 0.56 & 0.56 & \textbf{0.86} & 0.41 & 70 & 74 & 11 & 101 \\
Mistral-7B-Instruct-v0.2 & 0.52 & 0.54 & 0.83 & 0.38 & 65 & 72 & 13 & 106 \\
gemma-1.1-2b-it & 0.00 & 0.33 & 0.00 & 0.00 & 0 & \textbf{85} & \textbf{0} & 171 \\
\midrule
Random & 0.33 & 0.33 & 0.33 & 0.33 \\
\end{tabular}
\end{adjustbox}
\vspace{1em}
\caption{
Performance of \gls{llm} annotations with \textbf{game screen} observations, subgoals \textbf{suggested} by the \gls{llm}, and \textbf{no token separator}.
}
\vspace{-1em}
\label{tab:app:ablation-token-identify-gamescreen}
\end{table}

\begin{table}[h]
\centering
\begin{adjustbox}{max width=1\textwidth,center}
\begin{tabular}{lrrrrrrrr}
\begin{tabular}{lrrrrrrrr}
Annotator & F1 $\uparrow$ & Accuracy $\uparrow$ & Precision $\uparrow$ & Recall $\uparrow$ & TP $\uparrow$ & TN $\uparrow$ & FP $\downarrow$ & FN $\downarrow$ \\
\midrule
Human & \textbf{1.00} & \textbf{1.00} & \textbf{1.00} & \textbf{1.00} & \textbf{171} & \textbf{85} & \textbf{0} & \textbf{0} \\
\midrule
Meta-Llama-3-70B-Instruct & \textbf{0.83} & \textbf{0.74} & 0.74 & 0.94 & 161 & 29 & 56 & 10 \\
gemma-1.1-7b-it & \textbf{0.83} & 0.72 & 0.71 & \textbf{0.98} & \textbf{168} & 17 & 68 & \textbf{3} \\
Mixtral-8x7B-Instruct-v0.1 & 0.78 & 0.68 & 0.72 & 0.85 & 146 & 27 & 58 & 25 \\
c4ai-command-r-plus & 0.75 & 0.69 & \textbf{0.81} & 0.71 & 121 & 56 & 29 & 50 \\
Meta-Llama-3-8B-Instruct & 0.69 & 0.59 & 0.70 & 0.68 & 116 & 35 & 50 & 55 \\
c4ai-command-r-v01 & 0.66 & 0.57 & 0.70 & 0.63 & 107 & 39 & 46 & 64 \\
Mistral-7B-Instruct-v0.2 & 0.54 & 0.47 & 0.65 & 0.47 & 80 & 41 & 44 & 91 \\
gemma-1.1-2b-it & 0.00 & 0.33 & 0.00 & 0.00 & 0 & \textbf{85} & \textbf{0} & 171 \\
\midrule
Random & 0.33 & 0.33 & 0.33 & 0.33 \\
\end{tabular}
\end{tabular}
\end{adjustbox}
\vspace{1em}
\caption{
Performance of \gls{llm} annotations with \textbf{cropped} observations, subgoals \textbf{suggested} by the \gls{llm}, and \textbf{no token separator}.
}
\vspace{-1em}
\label{tab:app:ablation-token-identify-cropped}
\end{table}

\begin{figure}[t]
    \centering
    \hfill
    \begin{subfigure}[b]{\textwidth}
        \centering
        \includegraphics[width=0.95\linewidth]{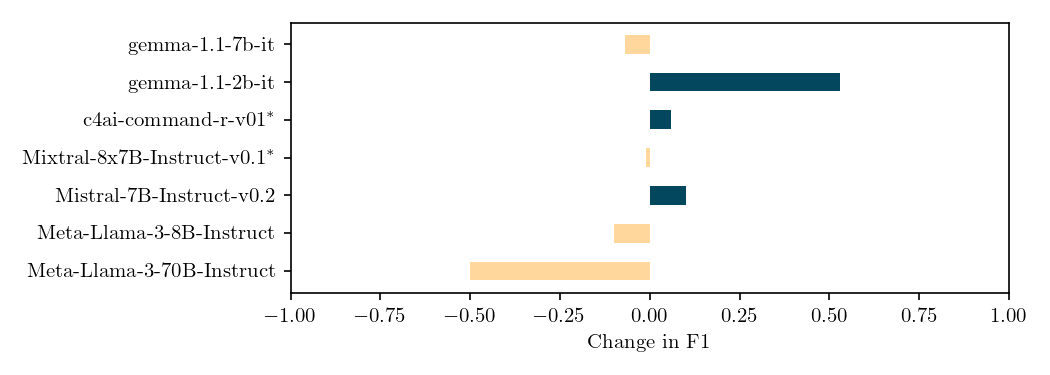}
        \caption{}
        \label{fig:ablation-token-preset-gamescreen}
    \end{subfigure}

    \begin{subfigure}[b]{\textwidth}
        \centering
        \includegraphics[width=0.95\linewidth]{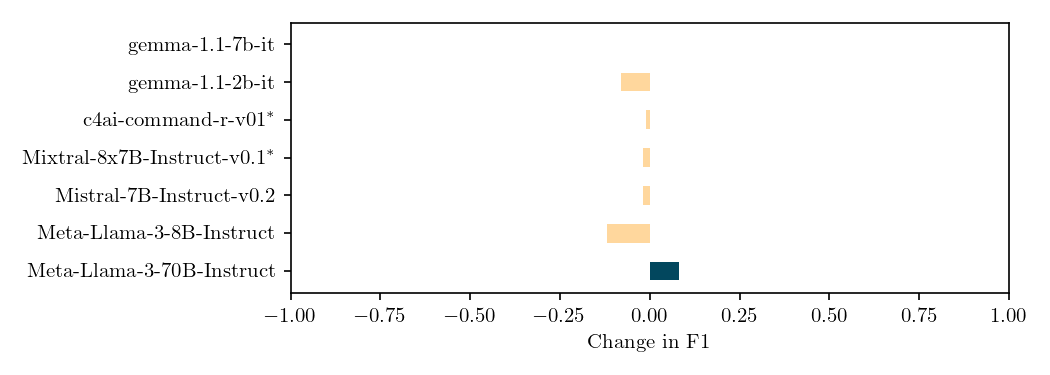}
        \caption{}
        \label{fig:ablation-token-preset-cropped}
    \end{subfigure}

    \begin{subfigure}[b]{\textwidth}
        \centering
        \includegraphics[width=0.95\linewidth]{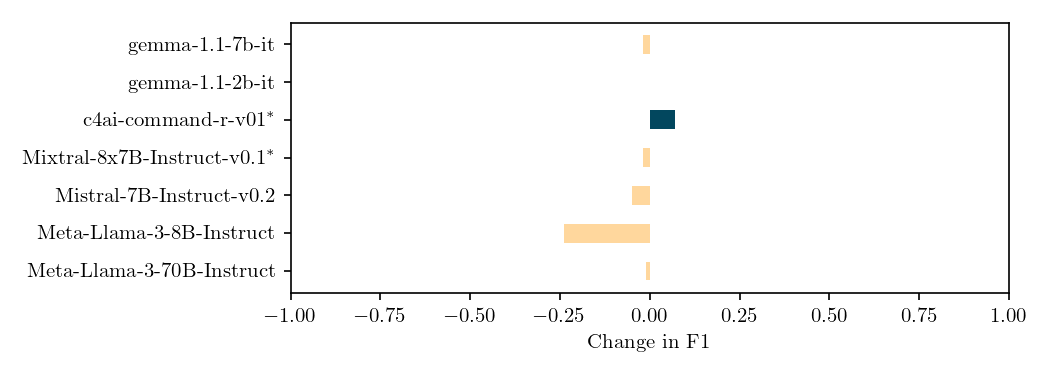}
        \caption{}
        \label{fig:ablation-token-identify-gamescreen}
    \end{subfigure}

    \begin{subfigure}[b]{\textwidth}
        \centering
        \includegraphics[width=0.95\linewidth]{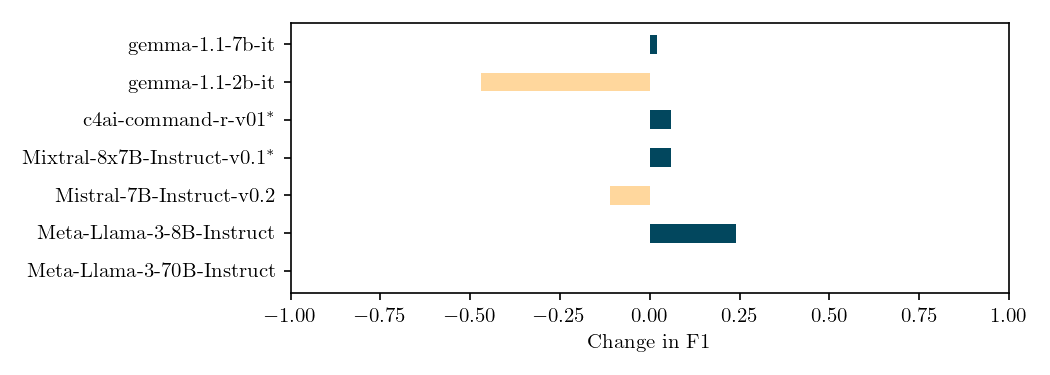}
        \caption{}
        \label{fig:ablation-token-identify-cropped}
    \end{subfigure}
    \caption{
    Variation in F1 score between the baseline results presented in Tables~\ref{tab:res:human-preset-balanced}-\ref{tab:res:crop-suggested-balanced} and the results without a token separator in Tables~\ref{tab:app:ablation-token-preset-gamescreen}-\ref{tab:app:ablation-token-identify-cropped}.
    Yellow bars indicate worse performance without a separator. and blue otherwise.
    }
\end{figure}

\begin{figure}[t]
    \centering
    \hfill
    \begin{subfigure}[b]{0.45\textwidth}
        \centering
        \includegraphics[width=0.95\linewidth]{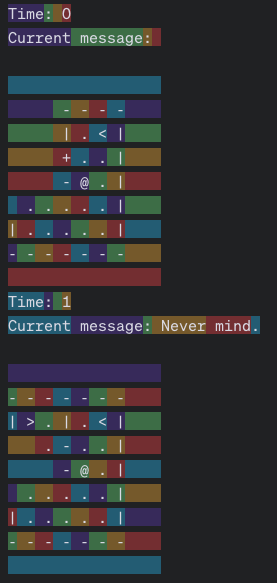}
        \caption{}
        \label{fig:tokenisation-with-sep}
    \end{subfigure}%
    \hfill%
    \begin{subfigure}[b]{0.45\textwidth}
        \centering
        \includegraphics[width=0.95\linewidth]{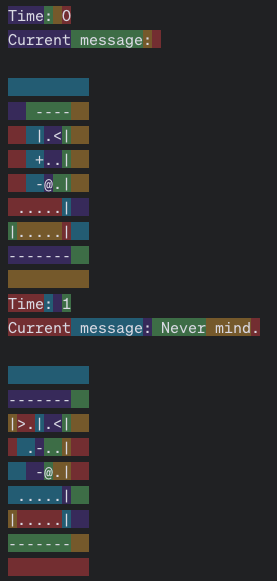}
        \caption{}
        \label{fig:tokenisation-without-sep}
    \end{subfigure}
    \hspace*{\fill}
    \caption{
    Tokenisation of the same prompt, with \textbf{(\subref{fig:tokenisation-with-sep})} and without \textbf{(\subref{fig:tokenisation-without-sep})} a token separator (whitespace).
    }
\end{figure}
\FloatBarrier

\subsection{Actions}
In this section we investigate the impacts of explicitly adding the action $a_t$ to the transition $(s_t, a_t, s_{t+1})$, which was left implicit in the main experiments in Section~\ref{sec:results}.

\begin{table}[H]
\centering
\begin{adjustbox}{max width=1\textwidth,center}
\begin{tabular}{lrrrrrrrr}
Annotator & F1 $\uparrow$ & Accuracy $\uparrow$ & Precision $\uparrow$ & Recall $\uparrow$ & TP $\uparrow$ & TN $\uparrow$ & FP $\downarrow$ & FN $\downarrow$ \\
\midrule
Human & \textbf{1.00} & \textbf{1.00} & \textbf{1.00} & \textbf{1.00} & \textbf{171} & \textbf{85} & \textbf{0} & \textbf{0} \\
\midrule
Mixtral-8x7B-Instruct-v0.1 & \textbf{0.83} & \textbf{0.79} & 0.89 & \textbf{0.77} & \textbf{132} & 69 & 16 & \textbf{39} \\
Mistral-7B-Instruct-v0.2 & 0.67 & 0.66 & 0.99 & 0.50 & 86 & 84 & 1 & 85 \\
gemma-1.1-7b-it & 0.66 & 0.66 & 0.97 & 0.50 & 86 & 82 & 3 & 85 \\
c4ai-command-r-v01 & 0.66 & 0.64 & 0.93 & 0.51 & 87 & 78 & 7 & 84 \\
Meta-Llama-3-8B-Instruct & 0.61 & 0.62 & 0.97 & 0.44 & 76 & 83 & 2 & 95 \\
Meta-Llama-3-70B-Instruct & 0.46 & 0.53 & 0.98 & 0.30 & 51 & 84 & 1 & 120 \\
gemma-1.1-2b-it & 0.00 & 0.33 & 0.00 & 0.00 & 0 & \textbf{85} & \textbf{0} & 171 \\
\midrule
\end{tabular}
\end{adjustbox}
\vspace{1em}
\caption{
Performance of \gls{llm} annotations with \textbf{game screen} observations, subgoals \textbf{provided} in the prompt, and the transition includes actions.
}
\vspace{-1em}
\label{tab:app:ablation-action-preset-gamescreen}
\end{table}
\begin{table}[H]
\centering
\begin{adjustbox}{max width=1\textwidth,center}
\begin{tabular}{lrrrrrrrr}
Annotator & F1 $\uparrow$ & Accuracy $\uparrow$ & Precision $\uparrow$ & Recall $\uparrow$ & TP $\uparrow$ & TN $\uparrow$ & FP $\downarrow$ & FN $\downarrow$ \\
\midrule
Human & \textbf{1.00} & \textbf{1.00} & \textbf{1.00} & \textbf{1.00} & \textbf{171} & \textbf{85} & \textbf{0} & \textbf{0} \\
\midrule
Mixtral-8x7B-Instruct-v0.1 & \textbf{0.80} & \textbf{0.75} & 0.86 & \textbf{0.75} & \textbf{128} & 64 & 21 & \textbf{43} \\
gemma-1.1-7b-it & 0.75 & 0.68 & 0.79 & 0.72 & 123 & 52 & 33 & 48 \\
c4ai-command-r-v01 & 0.72 & 0.68 & 0.86 & 0.62 & 106 & 68 & 17 & 65 \\
gemma-1.1-2b-it & 0.71 & 0.62 & 0.73 & 0.69 & 118 & 42 & 43 & 53 \\
Mistral-7B-Instruct-v0.2 & 0.66 & 0.66 & 0.98 & 0.50 & 86 & 83 & 2 & 85 \\
Meta-Llama-3-70B-Instruct & 0.64 & 0.59 & 0.78 & 0.54 & 92 & 59 & 26 & 79 \\
Meta-Llama-3-8B-Instruct & 0.50 & 0.54 & 0.90 & 0.35 & 60 & 78 & 7 & 111 \\
\midrule
\end{tabular}
\end{adjustbox}
\vspace{1em}
\caption{
Performance of \gls{llm} annotations with \textbf{cropped} observations, subgoals \textbf{provided} in the prompt, and the transition includes actions.
}
\vspace{-1em}
\label{tab:app:ablation-action-preset-cropped}
\end{table}
\begin{table}[H]
\centering
\begin{adjustbox}{max width=1\textwidth,center}
\begin{tabular}{lrrrrrrrr}
Annotator & F1 $\uparrow$ & Accuracy $\uparrow$ & Precision $\uparrow$ & Recall $\uparrow$ & TP $\uparrow$ & TN $\uparrow$ & FP $\downarrow$ & FN $\downarrow$ \\
\midrule
Human & \textbf{1.00} & \textbf{1.00} & \textbf{1.00} & \textbf{1.00} & \textbf{171} & \textbf{85} & \textbf{0} & \textbf{0} \\
\midrule
Meta-Llama-3-70B-Instruct & \textbf{0.85} & \textbf{0.79} & 0.80 & \textbf{0.91} & \textbf{155} & 46 & 39 & \textbf{16} \\
Mixtral-8x7B-Instruct-v0.1 & 0.79 & 0.69 & 0.72 & 0.86 & 147 & 29 & 56 & 24 \\
Meta-Llama-3-8B-Instruct & 0.78 & 0.66 & 0.68 & \textbf{0.91} & \textbf{155} & 13 & 72 & \textbf{16} \\
gemma-1.1-7b-it & 0.75 & 0.64 & 0.69 & 0.82 & 141 & 22 & 63 & 30 \\
Mistral-7B-Instruct-v0.2 & 0.71 & 0.61 & 0.70 & 0.71 & 122 & 33 & 52 & 49 \\
c4ai-command-r-v01 & 0.71 & 0.67 & \textbf{0.86} & 0.60 & 102 & 69 & 16 & 69 \\
gemma-1.1-2b-it & 0.00 & 0.33 & 0.00 & 0.00 & 0 & \textbf{85} & \textbf{0} & 171 \\
\midrule
\end{tabular}
\end{adjustbox}
\vspace{1em}
\caption{
Performance of \gls{llm} annotations with \textbf{game screen} observations, subgoals \textbf{suggested} by the \gls{llm}, and the transition includes actions.
}
\vspace{-1em}
\label{tab:app:ablation-action-identify-gamescreen}
\end{table}
\begin{table}[H]
\centering
\begin{adjustbox}{max width=1\textwidth,center}
\begin{tabular}{lrrrrrrrr}
Annotator & F1 $\uparrow$ & Accuracy $\uparrow$ & Precision $\uparrow$ & Recall $\uparrow$ & TP $\uparrow$ & TN $\uparrow$ & FP $\downarrow$ & FN $\downarrow$ \\
\midrule
Human & \textbf{1.00} & \textbf{1.00} & \textbf{1.00} & \textbf{1.00} & \textbf{171} & \textbf{85} & \textbf{0} & \textbf{0} \\
\midrule
Meta-Llama-3-70B-Instruct & \textbf{0.86} & \textbf{0.80} & 0.82 & 0.91 & 155 & 50 & 35 & 16 \\
Mixtral-8x7B-Instruct-v0.1 & 0.83 & 0.75 & 0.76 & 0.91 & 156 & 35 & 50 & 15 \\
gemma-1.1-7b-it & 0.82 & 0.72 & 0.73 & \textbf{0.92} & \textbf{158} & 27 & 58 & \textbf{13} \\
c4ai-command-r-v01 & 0.74 & 0.64 & 0.72 & 0.76 & 130 & 35 & 50 & 41 \\
Mistral-7B-Instruct-v0.2 & 0.72 & 0.63 & 0.74 & 0.70 & 120 & 42 & 43 & 51 \\
Meta-Llama-3-8B-Instruct & 0.53 & 0.46 & 0.63 & 0.46 & 79 & 39 & 46 & 92 \\
gemma-1.1-2b-it & 0.42 & 0.48 & 0.83 & 0.28 & 48 & \textbf{75} & \textbf{10} & 123 \\
\midrule
\end{tabular}
\end{adjustbox}
\vspace{1em}
\caption{
Performance of \gls{llm} annotations with \textbf{cropped} observations, subgoals \textbf{suggested} by the \gls{llm}, and the transition includes actions.
}
\vspace{-1em}
\label{tab:app:ablation-action-identify-cropped}
\end{table}

\begin{figure}[t]
    \centering
    \hfill
    \begin{subfigure}[b]{\textwidth}
        \centering
        \includegraphics[width=0.95\linewidth]{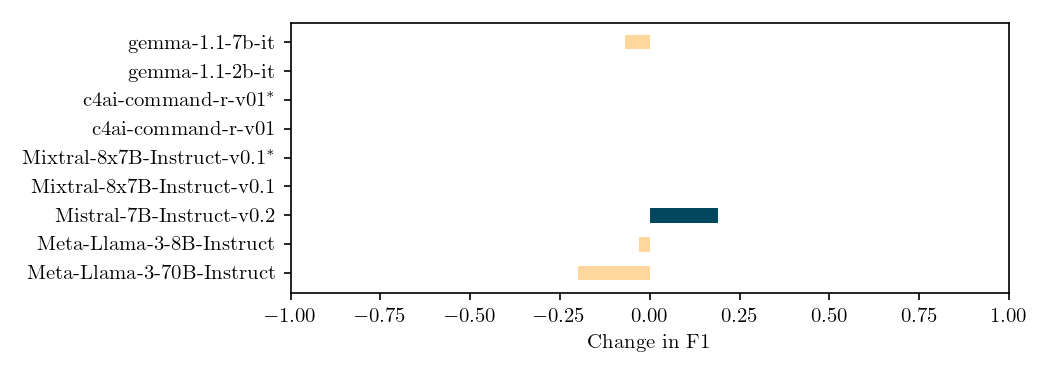}
        \caption{}
        \label{fig:ablation-action-preset-gamescreen}
    \end{subfigure}

    \begin{subfigure}[b]{\textwidth}
        \centering
        \includegraphics[width=0.95\linewidth]{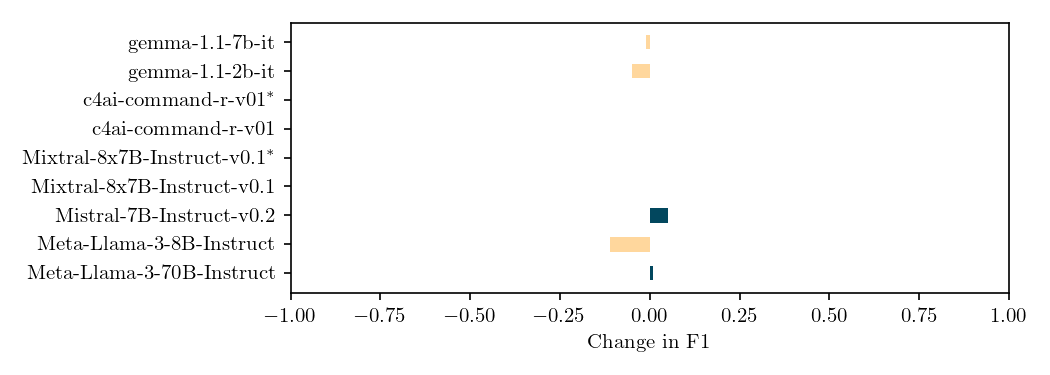}
        \caption{}
        \label{fig:ablation-action-preset-cropped}
    \end{subfigure}

    \begin{subfigure}[b]{\textwidth}
        \centering
        \includegraphics[width=0.95\linewidth]{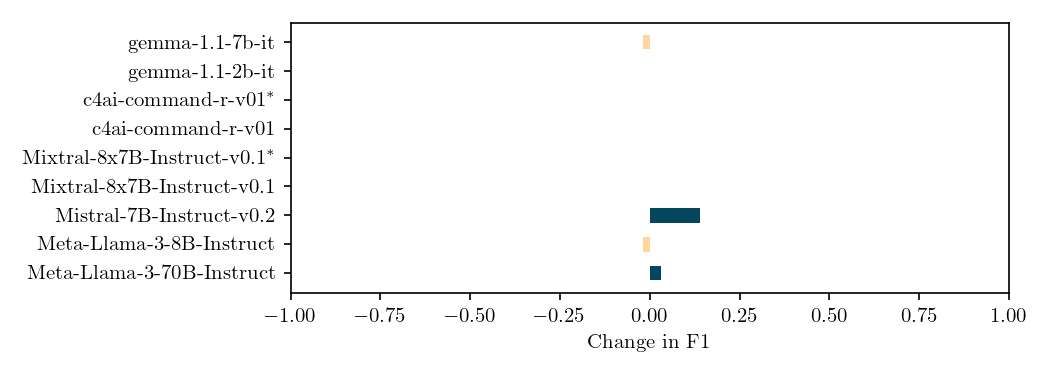}
        \caption{}
        \label{fig:ablation-action-identify-gamescreen}
    \end{subfigure}

    \begin{subfigure}[b]{\textwidth}
        \centering
        \includegraphics[width=0.95\linewidth]{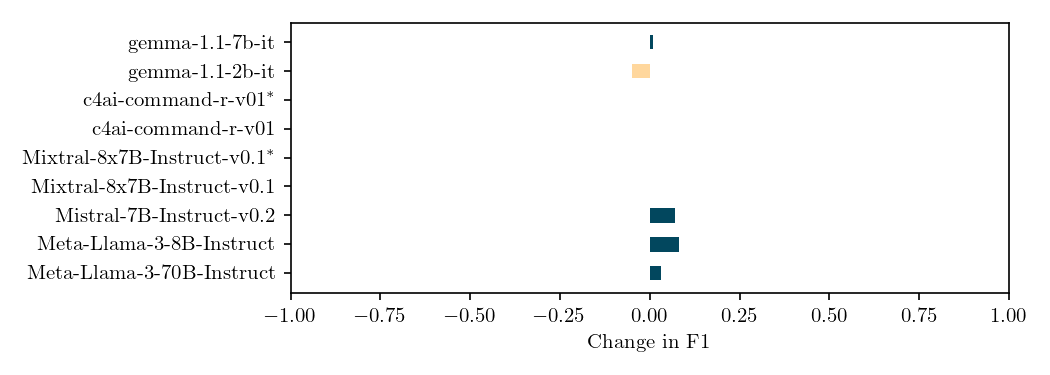}
        \caption{}
        \label{fig:ablation-action-identify-cropped}
    \end{subfigure}
    \caption{
    Variation in F1 score between the baseline results presented in Tables~\ref{tab:res:human-preset-balanced}-\ref{tab:res:crop-suggested-balanced} and the results where prompts also include the action in Tables~\ref{tab:app:ablation-action-preset-gamescreen}-\ref{tab:app:ablation-action-identify-cropped}.
    }
\end{figure}
\FloatBarrier

\end{document}